\documentclass[pdflatex,sn-mathphys-num]{sn-jnl}

\usepackage{graphicx}%
\usepackage{multirow}%
\usepackage{amsmath,amssymb,amsfonts}%
\usepackage{amsthm}%
\usepackage{mathrsfs}%
\usepackage[title]{appendix}%
\usepackage{xcolor}%
\usepackage{textcomp}%
\usepackage{manyfoot}%
\usepackage{booktabs}%
\usepackage{algorithm}%
\usepackage{algorithmicx}%
\usepackage{algpseudocode}%
\usepackage{listings}%
\usepackage{makecell}

\theoremstyle{thmstyleone}%
\theoremstyle{thmstyletwo}%

\theoremstyle{thmstylethree}%

\begin{document}

\title[Article Title]{From Paintbrush to Pixel: A Review of Deep Neural Networks in AI-Generated Art}


\author[1]{\fnm{Anne-Sofie} \sur{Maerten}}\email{annesofie.maerten@kuleuven.be}

\author*[2]{\fnm{Derya} \sur{Soydaner}}\email{d.soydaner@liacs.leidenuniv.nl}

\affil[1]{\orgdiv{Department of Brain and Cognition}, \orgname{University of Leuven (KU Leuven)}, \orgaddress{\city{Leuven}, \country{Belgium}}}

\affil*[2]{\orgdiv{Leiden Institute of Advanced Computer Science}, \orgname{Leiden University}, \orgaddress{\city{Leiden}, \country{The Netherlands}}}


\abstract{This paper delves into the fascinating field of AI-generated art and explores the various deep neural network architectures and models that have been utilized to create it. From the classic convolutional networks to the cutting-edge diffusion models, we examine the key players in the field. We explain the general structures and working principles of these neural networks. Then, we showcase examples of milestones, starting with the dreamy landscapes of DeepDream and moving on to the most recent developments, including Stable Diffusion and DALL-E 3, which produce mesmerizing images. We provide a detailed comparison of these models, highlighting their strengths and limitations, and examining the remarkable progress that deep neural networks have made so far in a short period of time. With a unique blend of technical explanations and insights into the current state of AI-generated art, this paper exemplifies how art and computer science interact.}

\keywords{Neural networks, Transformers, Diffusion models, Generative adversarial networks, Text-to-image models, Deep learning, Generative AI, Ethics}



\maketitle

\section{Introduction}

\textit{``AI-generated art is a new and emerging field, and it will likely take some time for it to fully mature and for its place in the art world to be fully understood,''} replied ChatGPT \cite{brown2020language}, the chatbot created by OpenAI, when we asked about the current state of AI-generated art. It continued, stating, \textit{``As with any new field, it will take time for it to be fully understood and for its place in the art world to be established.''} 

\begin{figure}[h] 
\centering
\includegraphics[width=4.0in]{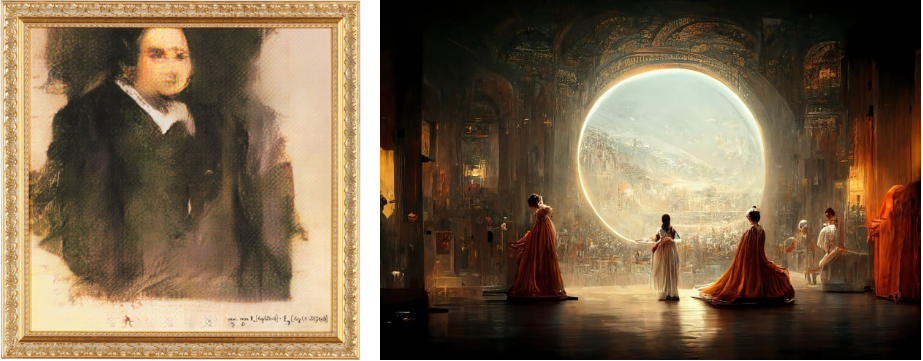}
\caption{\emph{(Left)} ``Edmond de Belamy'' - The first AI-generated portrait sold at Christie's art auction in 2018. \emph{(Right)} ``Thé\^{a}tre D'opéra Spatial'' - The winner of the digital art category at the Colorado State Fair's annual art competition in 2022.}
\label{edmond}
\end{figure}

Whether they are recognized as art or not, AI-generated images are widespread today. Regardless of discussions about how creative or artistic they are, their existence in our lives cannot be denied any longer. In 2018, the first AI-generated portrait \emph{``Edmond de Belamy''} (Fig.~\ref{edmond}) sold for \$432,500 at Christie's art auction. It was created using a generative adversarial network (GAN) \cite{goodfellow2014} as a part of \emph{``La Famille de Belamy''} series by Obvious Art. The fact that it is signed with the loss function used in GAN makes this case quite amusing. In 2022, Jason M. Allen's AI-generated artwork, \emph{``Thé\^{a}tre D'opéra Spatial''} (Fig.~\ref{edmond}), won the art prize in the digital category at the Colorado State Fair's annual art competition. This piece was created with the AI-based tool Midjourney\footnote{www.midjourney.com} which can generate images by taking text prompts as input.

The progress in AI-generated art is made possible by \emph{deep learning}, which is a subfield of machine learning. This subfield includes \emph{deep neural networks}, which have led to significant breakthroughs in various fields such as computer vision in the last decade. In this paper, we focus on deep neural networks used in image processing and recent developments that have been used to produce AI-generated images. Several studies address AI-generated art from different perspectives. For example, Cetinic \emph{et al.} (2022) \cite{cetinic2022} discussed the creativity of AI technologies, as well as authorship, copyright and ethical issues. \cite{ragot2020} conducted an experiment where they asked participants to rate the difference between paintings created by humans and AI in terms of liking, perceived beauty, novelty, and meaning. These are measured through statements such as ‘I like this painting,’ ‘This painting looks beautiful,’ ‘This painting seems novel,’ and ‘I perceive the meaning of the painting.’ They found that identical artworks were evaluated differently depending on whether they were attributed to a human or an AI, and that human-created artworks were generally rated more highly than AI-generated ones. Other important issues today are credit allocation and responsibility for AI-generated art \cite{epstein2020}. A review of generative AI models has been released which touches upon various applications, such as texts, images, videos and audios \cite{brizuela2023}. Recently, Wang \textit{et al.} (2025) \cite{wang2025diffusion} presented a review of diffusion models for visual art creation. In contrast, we provide a broader review of neural networks in image generation for artistic purposes, trace their temporal progress, present accessible technical explanations, and address important issues related to aesthetics and ethics. We include visual comparisons that make the technical discussion more tangible and offer practical guidance for readers selecting a model. 

In this paper, we focus on the main neural networks which have been used to generate realistic images. We first explain the building blocks of the related neural networks to provide readers with a better understanding of these models. We describe the general working principles of these neural networks, and introduce the recent trends in AI-generated art such as DALL-E 3 \cite{betker2023}. Then, we examine the rapid development of deep neural networks in AI-generated art, and emphasize the milestones in this progress. This review addresses the topic from a technical perspective and provides comparisons of current state-of-the-art models. However, even for a non-technical audience (e.g., from more traditional areas in the fine arts, aesthetics, and cultural studies), this review will provide an overview of the different techniques and tools that are currently available.  

The rest of the paper is organized as follows. In Section \ref{sec2}, we explain the important neural network types used for the models in AI-generated art. In Section \ref{sec3}, we provide an overview of the advancements in generative modeling, highlighting recent trends and breakthroughs in the field. In Section \ref{sec4}, we compare these deep generative models introduced in the previous section. Then, we discuss the ethical concerns in Section \ref{ethical_concerns}. We conclude in Section \ref{sec6}. 

\section{Preliminaries}\label{sec2}

During training, a neural network adjusts its parameters called \emph{weights}. When the training is completed, the weights are optimized for the given task, i.e., the neural network \emph{learns}.
A typical neural network is a multilayer perceptron (MLP) which is useful for classification and regression tasks. However, there are many deep neural networks which are particularly effective in image processing. One type is a convolutional neural network (CNN). In this section, we start with CNNs, which require data labels during training and learn in a supervised learning setting. Then, we explain the autoencoders that can learn without data labels, i.e., via unsupervised learning. We continue with generative models, including GANs and the Transformer neural network. Lastly, we explain diffusion models, which are the latest advancements in deep learning.

\subsection{Convolutional neural networks}\label{cnn}

Convolutional neural networks \cite{lecun1989, lecun1998}, usually referred to as CNNs, or ConvNets, are deep neural networks utilized mainly for image processing. In a deep CNN, increasingly more abstract features are extracted from the image through a series of hidden layers. As such, a CNN follows a similar hierarchy as the human visual cortex, in that earlier areas extract simple visual features and later areas extract combinations of features and high level features. In this manner, the complex mapping of the input pixels is divided into a series of nested mappings, each of which is represented by a different layer \cite{goodfellow2016}.

\begin{figure}[ht]
\centering
\includegraphics[width=5.0in]{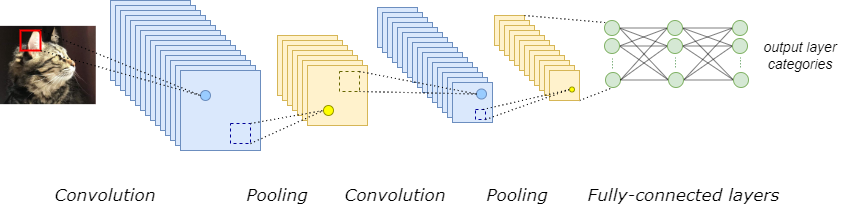}
\caption{An example CNN structure with two convolutional, two pooling, and three fully-connected layers for classification.}
\label{figcnn}
\end{figure}

There are various CNN architectures in the literature (e.g., LeNet \cite{lecun1989, lecun1998}, which is capable of recognizing handwritten digits). However, more complex tasks like object recognition require deeper CNNs such as AlexNet \cite{krizhevsky2012}, VGG \cite{simonyan2015}, ResNet \cite{he2016}, DenseNet \cite{huang2017}, EfficientNet \cite{tan2019}, MobileNet \cite{howard2017}, ResNeXt \cite{xie2017}, Inception and GoogLeNet \cite{szegedy2015}. A typical CNN includes three kinds of layers:  convolutional layers, pooling layers, and fully-connected layers. The general structure of a typical CNN for classification is illustrated in Fig. \ref{figcnn}. The input image $\mathcal{X}$ is presented at the input layer, which is followed by the first convolutional layer. In a convolutional layer, the weights are kept within \emph{kernels}. During learning, a mathematical operation called \emph{convolution} is performed between the input and the kernel. Basically, convolution is performed by multiplying the elements of the input with each element of the kernel and summing the results of this multiplication. This input can be an input image or the output of another, preceding convolutional layer. The units in a convolutional layer are arranged into planes, each of which is referred to as a \emph{feature map}, defined as 3D tensors. Each unit in a feature map takes input from a small area of the image, and all units detect the same pattern, but at different locations in the input image. All units in a feature map share the same weights, thus a convolution of the image pixel intensities with a \emph{kernel} is performed \cite{bishop2006}. In general, there are multiple feature maps in a convolutional layer, each with its own weights, to detect multiple features \cite{bishop2006}. Accordingly, a single convolutional layer includes many kernels, each operating at the same time. As a deep neural network, a CNN typically includes many convolutional layers, which implies millions of parameters.  

Each output of the convolution operation is usually run through a nonlinear activation function, such as the rectified linear unit (ReLU) \cite{glorot2011}. Then, a convolutional layer or a stack of convolutional layers is followed by a \emph{pooling layer} which reduces the size of the feature maps by calculating a summary statistic of the nearby outputs, such as the maximum value or the average \cite{goodfellow2016}. In the end, a series of convolutional and pooling layers is followed by \emph{fully-connected (dense)} layers. The activation function that the output layer applies depends on the task. In most cases, a sigmoid function is preferred for binary classification, softmax nonlinearity for multi-classification, and linear activation for regression tasks. Accordingly, a CNN minimizes the difference between the desired output values $y$ and the predicted values $\hat{y}$ in the cross-entropy loss functions given in Eq. \ref{eq-binary_crossentropy} and Eq. \ref{eq-crossentropyy} for binary and multi-class classification, respectively. In the case of a regression task, a CNN may minimize mean squared error given in Eq. \ref{eq-mse}. In the equations below, $W$ represents the weights belonging to the hidden layers, and $V$ represents the output layer weights. Although the aforementioned are the most frequently used activation and loss functions, there are other alternatives available such as Leaky ReLU \cite{maas2013}, SELU \cite{klambauer2017}, Swish \cite{ramachandran2017}, Mish \cite{misra2019} as activation functions, or mean absolute error as a loss function. 

\begin{equation}
\label{eq-binary_crossentropy}
L(W,V \mid \mathcal{X}) = -\sum\limits_t y^t log\hat{y}^t + (1-y^t)log(1-\hat{y}^t)
\end{equation}

\begin{equation}
\label{eq-crossentropyy}
L(W,V \mid \mathcal{X}) = -\sum\limits_t \sum\limits_i y_i^t log\hat{y}_i^t
\end{equation}

\begin{equation}
\label{eq-mse}
L(W,V \mid \mathcal{X})= \frac{1}{2} \sum\limits_t \sum\limits_i (y_i^t - \hat{y}_i^t)^2 
\end{equation}

A fully-connected layer and a convolutional layer significantly differ in that fully-connected layers learn global patterns in their input feature space, whereas convolutional layers learn local patterns \cite{chollet2018}. In a CNN, there are fewer weights than there would be if the network were fully-connected because of the local receptive fields \cite{bishop2006}. CNNs are now essential neural networks in deep learning and have yielded major advances for a variety of image processing tasks.

\subsection{Autoencoders}\label{autoencoders}
\begin{figure}[ht]
\centering
\includegraphics[width=5.0in]{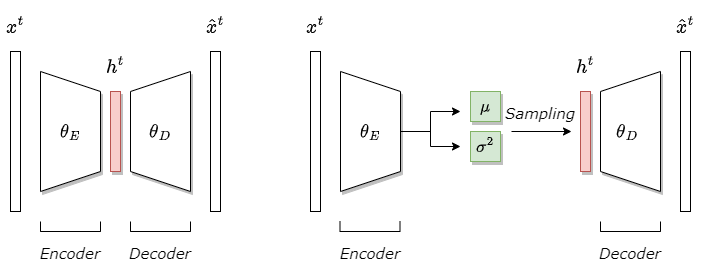}
\caption{The general structure of the \emph{(Left)} Autoencoder, \emph{(Right)} Variational autoencoder. $x^t$ refers to an input sample, $h^t$ to the latent representation and $\hat{x}^t$ to the reconstructed input.    The parameters of encoder $(\theta_E)$ and decoder $(\theta_D)$ are updated during training.}
\label{ae}
\end{figure}

The autoencoder, originally named the \emph{autoassociator} \cite{cottrell1987}, is a typical example of unsupervised learning. This neural network learns to reconstruct the input data 
by extracting a (usually lower-dimensional) representation of the data. The autoencoder has been successfully implemented for unsupervised or semi-supervised tasks, or as a preprocessing stage for supervised tasks. The general structure of an autoencoder consists of an \emph{encoder} and a \emph{decoder}, as shown in Fig. \ref{ae}. In the simplest case, both encoder and decoder are composed of fully-connected layer(s). During training, the encoder, in one or more layers, usually transforms the input to a lower-dimensional representation. Then, the decoder that follows, in one or more layers, takes this representation as input and reconstructs the original input back as its output \cite{soydaner2020}. The aim is to obtain a meaningful representation of data, which also makes the autoencoder an example of \emph{representation learning}. 

In the general framework, the encoder takes the input $x^t$, and produces a compressed or \emph{hidden}/\emph{latent} representation of input $h^t$. Then, the decoder takes $h^t$ as input and reconstructs the original input as the output $\hat{x}^t$. When the $h^t$ dimension is less than the $x^t$ dimension, the autoencoder is called \emph{undercomplete}. The undercomplete autoencoder can capture the most salient features of the data. On the other hand, when the $h^t$ dimension is equal to or greater than the $x^t$ dimension, it is called \emph{overcomplete}. The overcomplete autoencoder just copies the input to the output, as it cannot learn anything meaningful about the data distribution. Regularized autoencoders, such as the \emph{sparse autoencoders}, alleviate this issue by applying a regularization term in the loss function \cite{goodfellow2016}.    

The tasks related to image processing may require both encoder and decoder be composed of convolutional layers instead of fully-connected layers. In this case, it is called a \emph{convolutional autoencoder}. The first layers of the encoder are convolution/pooling layers and, correspondingly, the last layers of the decoder are deconvolution/unpooling layers. Whether the layers are fully-connected or convolutional, total reconstruction of a training set $\mathcal{X}=\{x^t\}_t$ is used as the loss function. The encoder and decoder weights, $\theta_E$ and $\theta_D$ respectively, are learned together to minimize this error:
\begin{eqnarray}
E(\theta_E,\theta_D \vert \mathcal{X}) &=& \sum_t \|x^t - \hat{x}^t\|^2
\nonumber\\
&=& \sum_t \| x^t - f_D(f_E(x^t\vert \theta_E)\vert \theta_D)\|^2 \label{ae_loss}
\end{eqnarray}

When an autoencoder in which the encoder and decoder are both perceptrons learns a lower-dimensional compressed representation of data, the autoencoder performs similar to principal component analysis (PCA): The encoder weights span the same space spanned by the \emph{k} highest eigenvectors of the input covariance matrix \cite{bourlard1988}. When the encoder and decoder are multilayer perceptrons, then the autoencoder learns to do nonlinear dimensionality reduction in the encoder \cite{soydaner2020}.    

Autoencoder architectures extend to specialized forms such as the \emph{denoising autoencoder} \cite{vincent2008}, which reconstructs clean data from corrupted inputs, and the \emph{contractive autoencoder} \cite{rifai2011}, which increases robustness to small changes in input data. While discussing the different autoencoder types, we should highlight the \emph{variational autoencoder (VAE)} \cite{kingma2014, rezende2014}, which turns an autoencoder into a \emph{generative} model. Similar to the autoencoder architecture mentioned above, a VAE is composed of an encoder and decoder. However, the encoder does not produce a lower-dimensional latent representation. Instead, the encoder produces parameters of a predefined distribution in the latent space for input, i.e., mean and variance. Thus, the input data is encoded as a probability distribution. Then, new samples can be generated by sampling from the latent space that the autoencoder learned during training. There are variants of this neural network architecture, including VQ-VAE \cite{oord2017}.

Unlike Eq. \ref{ae_loss}, the loss function of VAE is composed of two main terms. The first is the reconstruction loss, which is the same loss as Eq. \ref{ae_loss}. The second term is the Kullback-Leibler divergence between the latent space distribution and standard Gaussian distribution. The loss function is the sum of these two terms. Once the VAE is trained, new samples can be generated by using the learned latent space. This property makes the VAE a generative model. 

\subsection{Generative adversarial networks}\label{gan}
The \emph{generative adversarial network} (GAN) \cite{goodfellow2014} is a milestone for generative models in deep learning literature.  The idea is based on game theory and a minimax game with two players. Instead of human players, in this case, the players are neural networks. One of these neural networks is called a \emph{generator (G)} while the other is called a \emph{discriminator (D)}. These two networks are trained end-to-end with backpropagation in an \emph{adversarial} setting, i.e., the generator and discriminator compete with each other. The generator captures the data distribution while the discriminator estimates the probability that its input either comes from the real data or is a fake sample created by the generator (Fig.~\ref{gan}). The competition between these two neural networks in the game makes them improve their results until the fake data generated by G is indistinguishable from the original data for D. As a result, G learns to model the data distribution during training and can generate \emph{novel} samples after training is completed.  

\begin{figure}[ht] 
\centering
\includegraphics[width=5.0in]{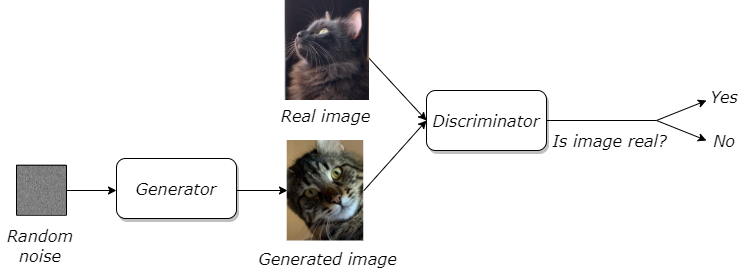}
\caption{The general structure of a generative adversarial network (GAN). The generator upscales its input (a noise vector) through a series of layers into an image. The discriminator performs a binary classification task, i.e., deciding whether the input image it receives is real or a generated sample.}
\label{gan}
\end{figure} 

In the original GAN framework, both the generator and discriminator are MLPs. The generator takes a random noise vector as input and generates samples. The discriminator takes both the generated sample and a real sample from the data as inputs, and tries to decide which one is real. Then, based on the feedback coming from the discriminator, the generator updates its weights. As shown in the loss function in Eq. \ref{eq-gan}, G tries maximizing the probability of D making a mistake. Ideally, the game ends at a saddle point where D is equal to 1/2 everywhere. In practice, GAN training faces several challenges \cite{khanuja2023}. As a solution, GANs that use different loss functions have been proposed, such as the Wasserstein GAN (WGAN) \cite{arjovsky2017}.

\begin{align}
\label{eq-gan}
\min_{G}\max_{D} V(D,G) = \mathbb{E}_{x\sim p_{\text{data}}(x)}[\log{D(x)}] + \nonumber \\
\mathbb{E}_{z\sim p_{\text{z}}(z)}[\log(1 - {D(G(z)))}]
\end{align}

In the literature, there are various GAN architectures \cite{creswell2018}. For example, when the G and D are composed of convolutional layers, this architecture is called a \emph{deep convolutional GAN (DCGAN)} \cite{radford2016}. \emph{StyleGAN} \cite{karras2019} is a convolutional GAN which can vary coarse-to-fine visual features separately. Whereas an ordinary GAN receives a noise vector as input, StyleGAN inputs the noise vector to a mapping network to create an intermediate latent space. The intermediate latents are then fed through the generator through adaptive instance normalization \cite{karras2019}. The mapping network ensures that features are disentangled in latent space, allowing StyleGAN to manipulate specific features in images (e.g., remove someone's glasses or make someone look older). Pix2pix \cite{isola2017} is a GAN that takes an image as input (rather than noise) and translates it to a different modality (e.g., BW image to RGB image). Whereas the training of pix2pix requires image pairs (the same image in RGB and BW), CycleGAN \cite{zhu2017} alleviates this need by ‘cycling’ between two GANs (see Section~\ref{subsectiongans}). An \emph{adversarial autoencoder} \cite{makhzani2015} combines the adversarial setting of GANs with the autoencoder architecture. Lastly, a \emph{self-attention adversarial neural network (SAGAN)} \cite{zhang2019} defines a GAN framework with an attention mechanism, which is explained in Section \ref{transformers}.

\subsection{Transformers}\label{transformers}
Convolution has become the central component for image processing applications as neural networks progressed throughout time. However, the human visual system has evolved to be sparse and efficient. Therefore, we do not process our entire visual field with the same resources, unlike convolutions. Rather, our eyes perform a fixation point strategy by means of a \emph{visual attention system} which plays an essential role in human cognition \cite{noton1971,noton1971a}. Inspired by the attention system in human vision, \emph{computational} attention mechanisms have been developed and integrated into neural networks. One of the goals is reducing the computational burden caused by the convolution operation, while still improving the performance. 

\begin{figure}[ht]
\centering
\includegraphics[width=3.0in]{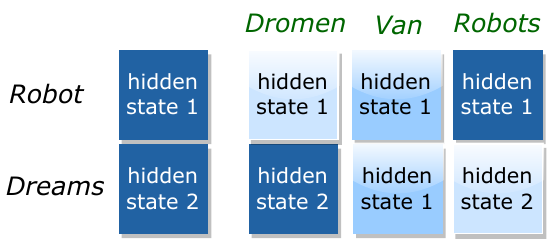}
\caption{A neural machine translation example. The model takes an English sentence as input and translates it into Dutch. The figure shows encoder hidden states, and which words the model focuses more on (indicated by the color intensity) while translating.}
\label{nlp_example}
\end{figure}

These attention-based neural networks have been applied to a variety of applications such as image recognition or object tracking; see \cite{soydaner2022} for a review. In particular, a novel attention-based encoder-decoder neural network architecture was presented for neural machine translation (NMT) in 2015 \cite{bahdanau2015}. The idea behind this approach is illustrated in Fig.~\ref{nlp_example}, which shows how the attention mechanisms in neural networks work. In this example, the encoder takes an input sentence in English, and the decoder outputs its translation in Dutch. Both encoder and decoder include \emph{recurrent neural networks (RNNs)}; see \cite{graves2012} for more information about RNNs. Basically, the encoder outputs hidden states, and the decoder takes all the hidden states as inputs. Before processing them, the decoder applies an attention mechanism that gives each hidden state a score. Then, it multiplies each hidden state by its score to which a softmax function is applied. The weighted scores are summed, and the result leads to the context vector for the decoder. By obtaining weighted hidden states which are most associated with certain words, the decoder focuses on the relevant parts of the input during decoding.

\begin{figure}[ht]
\centering
\includegraphics[width=4.3in]{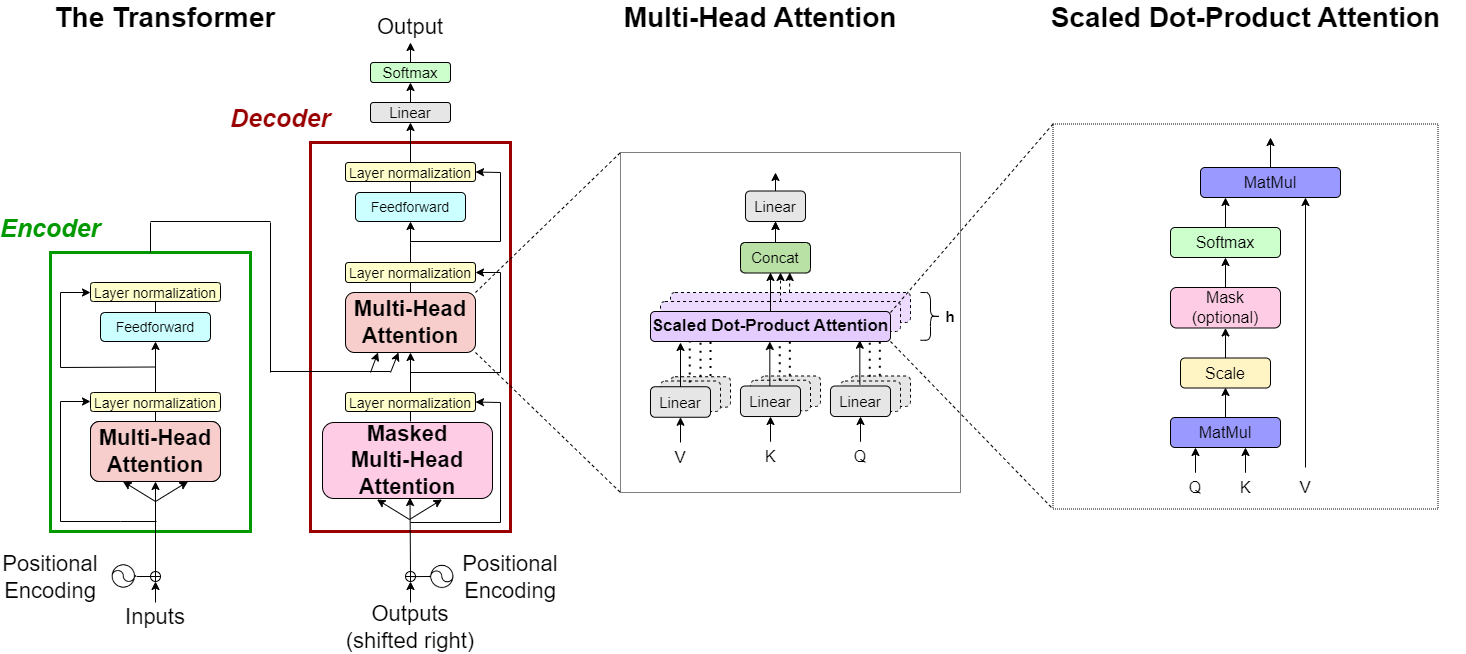}
\caption{The Transformer architecture in detail \cite{vaswani2017, soydaner2022}. \emph{(Left)} The Transformer with one encoder-decoder stack. \emph{(Center)} Multi-head attention. \emph{(Right)} Scaled dot-product attention.}
\label{transformer}
\end{figure}

Attention mechanisms in neural networks have progressed rapidly, especially for NMT. One is the \emph{self-attention mechanism}, which is the core building block of the \emph{Transformer} \cite{vaswani2017}. The original Transformer is composed of encoder-decoder stacks which are entirely based on self-attention without any convolutional or recurrent layers. There are six identical layers in each of the encoder-decoder stacks that form the Transformer. To illustrate the model, only one encoder-decoder stack is shown in Fig.~\ref{transformer}. 

The encoder-decoder stacks in the Transformer are composed of fully-connected layers and \emph{multi-head attention} which is a kind of self-attention mechanism applying \emph{scaled dot-product attention} within itself. As seen in Fig.~\ref{transformer}, these attention mechanisms use three vectors for each word, namely \emph{query (Q)}, \emph{key (K)} and \emph{value (V)}. These vectors are computed by multiplying the input with weight matrices $W_q$, $W_k$ and $W_v$ which are learned during training. In general, each value is weighted by a function of
the query with the corresponding key. The output is computed as a weighted sum of the values. In the \emph{scaled dot-product attention}, the dot products of the query with all keys are computed. As given in Eq. \ref{eq-attention}, each result is divided by the square root of the dimension of the keys (to have more stable gradients) and passed onto the softmax function. Finally, each softmax score is multiplied with the value \cite{vaswani2017}: 

\begin{equation}
Attention(Q,K,V) = softmax(\frac{QK^T}{\sqrt{d_k}})V
\label{eq-attention}
\end{equation}

\emph{Multi-head attention} extends this idea by applying linear activations to the inputs (keys, values and queries) \emph{h} times based on different, learned linear representations (Fig.~\ref{transformer}). Each of the projected versions of queries, keys and values are called \emph{heads}, for which the scaled dot-product attention is performed in parallel. Thus, the self-attention is calculated multiple times using different sets of query, key and value vectors. This allows the model to jointly attend to information at different positions \cite{vaswani2017}. In the last step, the projections are concatenated. Additionally, the decoder applies \emph{masked multi-head attention} to ensure that only previous word embeddings (tokens) are used when predicting the next word in the sentence.    

\begin{figure}[ht]
\centering
\includegraphics[width=4.5in]{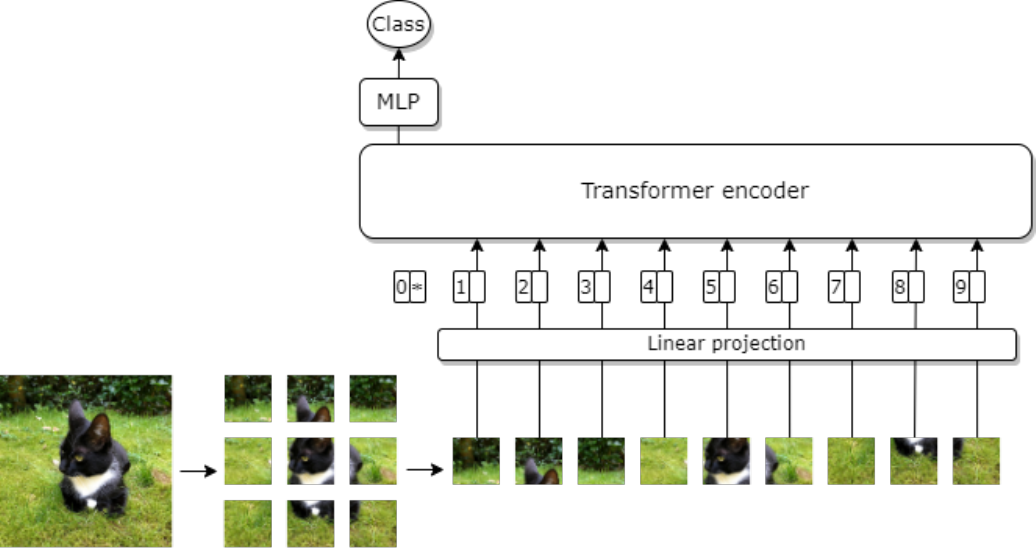}
\caption{The Vision Transformer \cite{dosovitskiy2020}. In order to classify an image, it takes the input as patches, projects linearly, adds position embeddings, and uses a Transformer encoder.}
\label{vit}
\end{figure}

There are different Transformer architectures for various tasks  \cite{khan2022, lin2022}. After the Transformer yielded major progress in natural language processing (NLP), it has been adapted to image processing tasks. \emph{Image Transformer} \cite{parmar2018} is one of these adaptations in which the Transformer is applied to image processing. Image Transformer applies self-attention in local neighborhoods for each query pixel, and performs well on image generation and image super-resolution. However, the current state-of-the-art model is \emph{Vision Transformer} (ViT)  \cite{dosovitskiy2020}, which splits an input image into patches before the Transformer takes the linear embeddings of these patches in sequence as input (Fig.~\ref{vit}). Vision Transformer performs well on image classification tasks whilst using fewer computational resources. Recent models, such as \emph{Swin transformer} \cite{liu2021}, \emph{DeiT} \cite{touvron2021}, \emph{BEiT} \cite{bao2021}, and
\emph{DINOv2} \cite{oquab2024}, have built on ViT and have been successful in image processing tasks.

\subsection{Diffusion models}\label{diffusion}
Today, text-to-image models such as DALL-E 3 \cite{betker2023} or Midjourney have turned AI into a popular tool to produce mesmerizing images. These are \emph{diffusion models} which have shown great success in generating high-quality images. They have shown to outperform GANs at image synthesis \cite{dhariwal2021}.    

Unlike GANs, the training of diffusion models does not require an adversarial setting. The original Denoising Diffusion method \cite{dickstein2015} was inspired by non-equilibrium thermodynamics that systematically destroys structure in a data distribution, then restores the data. Based on this method, \emph{denoising diffusion probabilistic models} or \emph{diffusion models} in short, have been applied to image generation \cite{ho2020}.  

\begin{figure}[ht]
\centering
\includegraphics[width=5.0in]{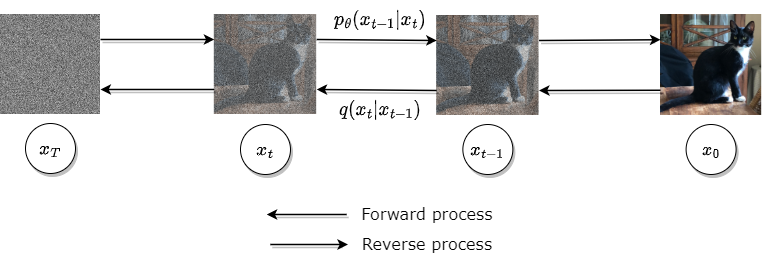}
\caption{The working principle of diffusion model in general \cite{ho2020}. Starting from an image, the forward process involves gradually adding noise (following a fixed Markov chain) until the image is all noise. In the reverse process, the original image is reconstructed step by step through a learned Markov Chain.}
\label{diff}
\end{figure}

Diffusion models require two main steps in the training phase (Fig.~\ref{diff}). At first, during \emph{the forward (diffusion) process}, random noise is gradually added to the input image until the original input becomes all noise. This is performed by a \emph{fixed} Markov chain which adds Gaussian noise for \emph{T} successive steps. Secondly, during the \emph{reconstruction} or \emph{reverse process}, the model reconstructs the original data from the noise obtained in the forward  process. The reverse process is defined as a Markov Chain with \emph{learned} Gaussian transitions. Accordingly, the prediction of probability density at time \emph{t} depends only on the state attained at time \emph{(t-1)}. Here, $x_1,...,x_T$ are latents of the same dimensionality as the data which make the diffusion models \emph{latent variable models} \cite{ho2020}.

\begin{figure}[ht]
\centering
\includegraphics[width=4.0in]{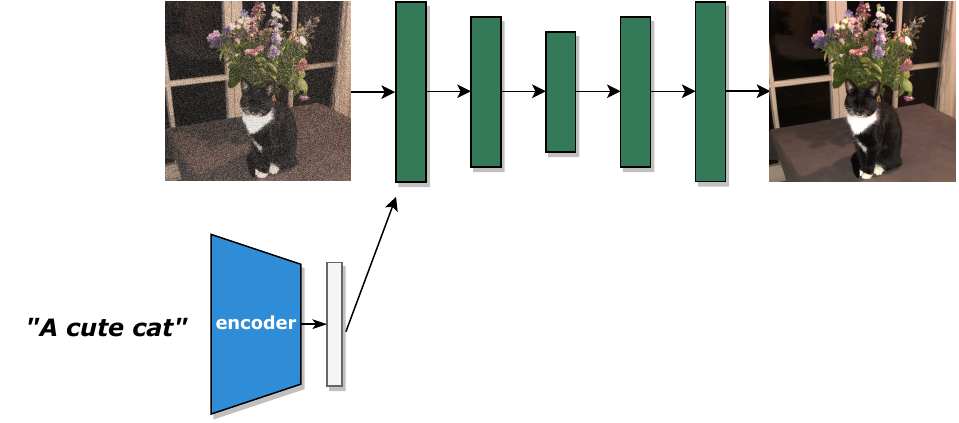}
\caption{The illustration of one time step in the learned Markov Chain in the reverse process. A deep neural network (e.g., U-Net \cite{ronneberger2015}) learns to transform a noisy input into a less noisy image with the help of a text prompt that describes the content of the image.}
\label{diff2}
\end{figure}

The general structure of a diffusion model is given in Fig.~\ref{diff}. Estimating the probability density at an earlier time step, given the current state of the system, is non trivial, so the reverse process requires training a neural network. To this end, all previous gradients are necessary for obtaining the required estimation. For each step in the Markov Chain, a neural network learns to denoise the image. Optionally, the denoising process can be guided by text (see Section~\ref{diffusion_models}). In this case, a Transformer encoder maps a text prompt to tokens which are then fed to the neural network (Fig.~\ref{diff2}). Once trained, a diffusion model can be used to generate data by simply passing random noise (and optionally a text prompt) through the learned denoising process. It should be noted that this process requires multiple denoising steps at inference time, making diffusion models slow in generating images when prompted. As a result, several methods have been proposed to speed up inference from multi-step to few-step and even single-step \cite{song2022, dockhorn2022, zhang2023, shaul2023, salimans2022, meng2023, luo2023, song2023, yin2024, xu2024, sauer2023, lin2024, schmidhuber2020}. The current SOTA method for single-step and few-step inference is \emph{Adversarial Diffusion Distillation} \cite{sauer2023}, which leverages adversarial training to push the image generation to sensible images with fewer denoising steps.

The diffusion model may bring to mind VAEs which encode the input data as a probability distribution and then sample from the learned latent space (Section~\ref{autoencoders}). However,   
the forward process makes the diffusion model different from the VAEs as training is not required in this fixed Markov chain. More detailed information about diffusion models and its applications can be found in  \cite{yang2022}.

\section{AI-generated art}\label{sec3}
In this section, we provide an overview of the models that have shaped the field of AI-generated art. This overview includes GANs, Transformer-based models and Diffusion models. In each sub-section, we touch on the models that had a large impact on the field, elaborating on the model architectures and providing a comprehensive comparison. It should be noted that this review is focused on models which have been detailed in scientific papers and therefore does not include the well-known diffusion model Midjourney. 

\subsection{Opening Gambit}

{\bf{DeepDream.}} Once CNNs achieved impressive results in image processing, researchers started developing visualization techniques to better understand how these neural networks see the world and perform classification. Examining each layer of a trained neural network led to the development of \emph{DeepDream} \cite{mordvintsev2015, mordvintsev2015a}, which produces surprisingly artistic images.

DeepDream generates images based on the representations learned by the neural network. It takes an input image, runs a trained CNN in reverse, and tries to maximize the activation of entire layers by applying gradient \emph{ascent} \cite{chollet2018}. DeepDream can be applied to any layer of a trained CNN. However, applying it to high-level layers is usually preferred because it provides visual patterns such as shapes or objects that are easier to recognize.

\begin{figure*}[ht]
\centering
\includegraphics[width=5.0in]{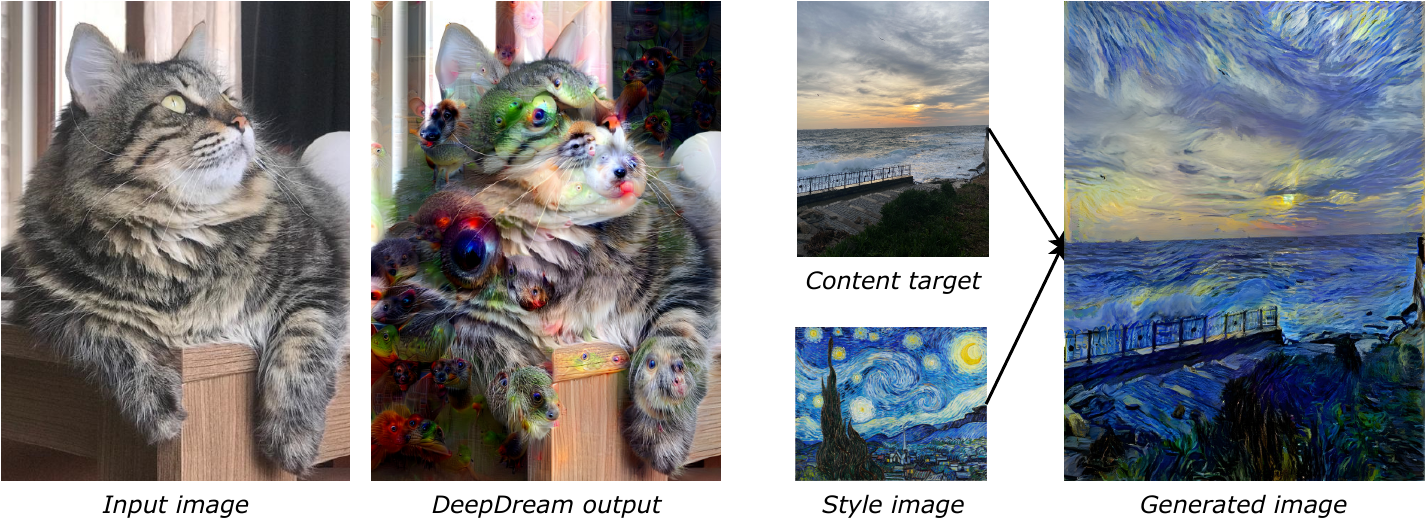}
\caption{\emph{(Left)} A DeepDream example. Deepdream receives an input image and outputs a dreamy version in which faces and eyes emerge due to the maximization of the final layer's activations.    \emph{(Right)}  An illustration of neural style transfer. The content target image and style image are provided as input to the model. As output, it retains the content of the target image and applies the style of the other image.}
\label{dream_nst}
\end{figure*}

An illustration of an original input image and its DeepDream output is shown in Fig.~\ref{dream_nst}. Strikingly, the output image contains many animal faces and eyes. This is due to the fact that the original DeepDream is based on Inception \cite{szegedy2015}, which was trained using the ImageNet database \cite{deng2009}. Since there are so many examples of different dog breeds and bird species in the ImageNet database, DeepDream is biased towards those. For some people, DeepDream images resemble dream-like psychedelic experiences. In any case, although it was not its initial purpose, DeepDream inspired people to employ AI as a tool for artistic image creation.

\noindent{\bf{Neural Style Transfer.}} A deep learning-based technique to combine the content of an image with the style of another image is called \emph{neural style transfer} \cite{gatys2015}. This technique uses a pretrained CNN to transfer the style of one image to another image. A typical example is shown in Fig.~\ref{dream_nst} where the style of one image (e.g., \emph{Starry Night} by Vincent Van Gogh) is applied to a content target image (e.g., a nice beach photo). Neural style transfer can be implemented by redefining the loss function in a CNN. This loss function is altered to preserve the content of the target image through the high-level activations of the CNN. At the same time, the loss function should capture the style of the other image through the activations in multiple layers. To this end, similar correlations within activations for low-level and high-level layers contribute to the loss function \cite{chollet2018}. The result is an image that combines the content of the first input image with the style of the second input image.

\subsection{Generative Adversarial Networks}\label{subsectiongans}
{\bf{ArtGAN.}} \cite{tan2017} presented their model called \emph{ArtGAN} in 2017, in which they show the result of a GAN trained on paintings. Although their output images looked nothing like an artwork of one of the great masters, the images seemed to capture the low-level features of artworks. This work sparked interest in the use of GANs to generate artistic images. Additionally, it challenged people to think of creativity as a concept that could be replicated by artificial intelligence. 

{\bf{CAN.}} Shortly afterwards, Elgammal \emph{et al.} (2017) \cite{elgammal2017} further pushed this idea in their paper on \emph{creative adversarial neural networks (CAN)}. Their goal was to train a GAN to generate images that would be considered as art by the discriminator but did not fit any existing art styles. The resulting images looked mostly like abstract paintings that had a unique feel to them. They then validated their work in a perceptual study where they asked human participants to rate the images on factors such as liking, novelty, ambiguity, surprisingness and complexity. In addition, they asked the participants whether they thought the image originated from a human-made painting or a computer-generated one. There were no differences in the scores on the above mentioned factors between CAN art and various abstract artworks or more traditional GAN art. However, participants more often thought that their CAN images were made by a human artist as opposed to GAN generated art.

\begin{figure}[ht]
\centering
\includegraphics[width=3.0in]{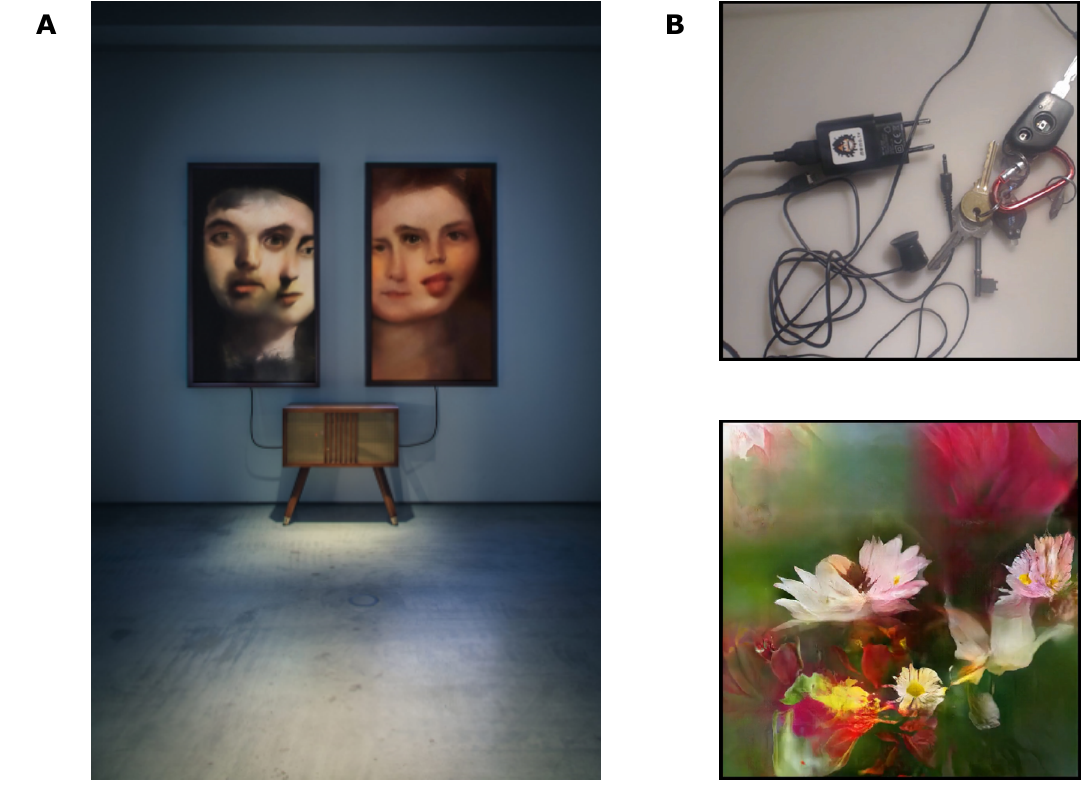}
\caption{Examples of artistic applications of pix2pix. \textbf{A.} Artwork \emph{“Memories of Passersby I”} by Mario Klingemann. This work continuously generates a male and female looking portrait by manipulating previously generated portraits using a collection of GANs (including pix2pix). \textbf{B.} Screenshot from video \emph{“Learning to See: Gloomy Sunday”} by Memo Akten. The original video shows a side by side of the input and output of pix2pix trained to turn ordinary video (showing household items) into artistic landscapes and flower paintings.}
\label{pix2pix}
\end{figure}

{\bf{pix2pix.}} In 2017, \cite{isola2017} had the innovative idea to create a conditional GAN \cite{mirza2014} that receives an image as input and generates a transformed version of that image. This was achieved by training the GAN on corresponding image pairs. For example, say you have a dataset of RGB images. You can create a BW version of all these images to create image pairs (one is RGB, one is BW). What is not as trivial, however, is turning BW images into colored ones. One could manually color the images, but this is time consuming. \emph{pix2pix} allows you to automate this process. The generator receives the BW version of the image pair and generates an RGB version. Next, the discriminator receives both the transformed image and the original RGB image, and has to determine which one is real and which one is fake. After the training is completed, the pix2pix GAN can transform any BW image into a colored version. The major advantage of pix2pix is that it can be applied to any dataset of image pairs without the need to adjust the training process or loss function. The same model can be used to transform sketches into paintings or BW images into colored images, simply by changing the training dataset. Many artists, as well as AI enthusiasts, have been inspired by pix2pix to create artistic images using this model (Fig.~\ref{pix2pix}). 

{\bf{CycleGAN.}} Although pix2pix was a major breakthrough in generative AI, one shortcoming was that it requires corresponding image pairs for its training, which simply do not exist for all applications. For example, we do not have a corresponding photograph for every painting created by Monet. Therefore, pix2pix would not be able to turn your photograph into a Monet painting. In 2017, the same lab released \emph{CycleGAN} \cite{zhu2017}, another major breakthrough in generative AI, since this GAN is able to transform your photograph into a Monet painting. CycleGAN extends pix2pix by combining two conditional GANs and ‘cycling’ between them. The first GAN’s generator might receive an image of a Monet painting and is trained to transform it to a photograph. The output image is then fed to the second generator to be transformed into a Monet painting. This transformed Monet painting and the original Monet painting are then fed to the first discriminator, whereas the photograph version of the image is compared to an existing (unpaired) photograph by the second discriminator. The same process is repeated for an existing photo by turning it into a Monet painting and back to a photo. The transformed photo is then compared to the original photo whereas the transformed Monet painting is compared to an existing (unpaired) Monet painting. In the end, the model can transform images into the other modality without having seen pairs in the training set.

\begin{figure}[ht]
\centering
\includegraphics[width=3.0in]{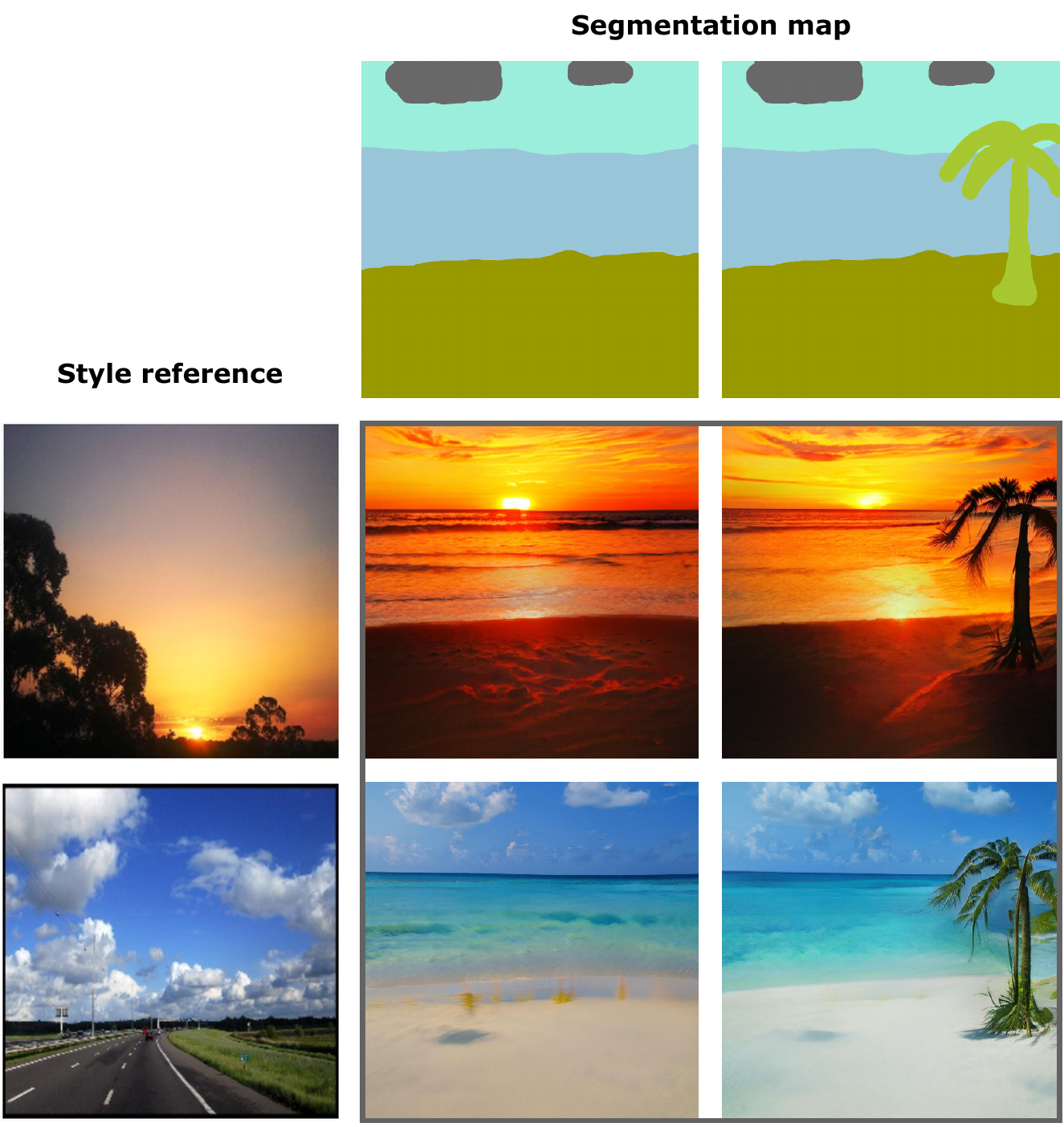}
\caption{Example images generated with GauGAN. One can provide a segmentation map and optionally a style reference as input. GauGAN then generates a photo-realistic version of the segmentation map in the style of the reference image. When we add a palm tree in the segmentation map, GauGAN adds a palm tree to its output in two different styles.}
\label{gaugan}
\end{figure}

{\bf{GauGAN.}} In 2019, Nvidia researchers released \emph{GauGAN} \cite{park2019}, named after post-impressionist painter Paul Gaugain. Similar to pix2pix, GauGAN takes an image as input and produces a transformation of that image as output. Their model uses spatially adaptive denormalization, also known as SPADE, which is a channel-wise normalization method (similar to Batch Normalization) convolved with a segmentation map. As a result, the output image can be steered with a semantic segmentation map. In addition, the input image is decoded using a VAE, which learns a latent distribution capturing the style of the image. As a result, one can generate new images of which the content is controlled by a segmentation map and the style by an existing image. Fig.~\ref{gaugan} shows some example images generated with GauGAN.

Nvidia has since released an updated version called \emph{GauGAN2} \cite{salian2021}. As the name suggests, it is still a GAN framework. However, this updated version can additionally perform text-to-image generation, meaning it can generate an image based on a text description as input. Earlier that year, text-to-image models became extremely popular due to the release of DALL-E \cite{ramesh2021}, a Transformer-based model which will be discussed in the next section.

{\bf{LAFITE.}} In 2021, \cite{zhou2021} proposed a GAN-based framework to perform language-free text-to-image generation, meaning they train their model solely on images. However, it is still able to generate images based on text descriptions after training is completed. They use CLIP's \cite{radford2021} joint semantic embedding space of text and images to generate pseudo text and image pairs. CLIP is another model that is trained to link text descriptions to the correct image and vice versa. Then, they adapt StyleGAN 2 \cite{karras2020} to a conditional version where the text embedding is concatenated with StyleSpace, the intermediate and well-disentangled feature space of StyleGAN \cite{karras2019}.

\subsection{Text-to-Image Models}

\begin{figure*}[!t]
\centering
\includegraphics[width=5.5in]{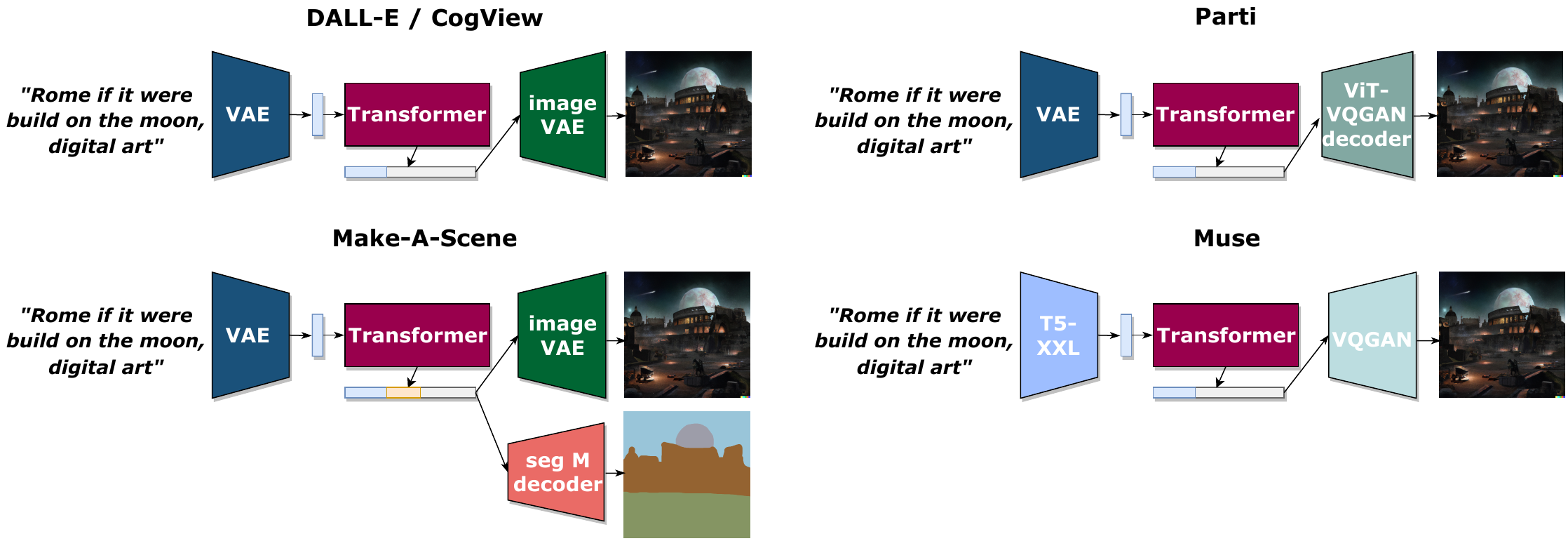}
\caption{Comparison of the Transformer-based text-to-image models.}
\label{comparisonTransformers}
\end{figure*}

\subsubsection{The Transformer}
\begin{figure}[ht]
\centering
\includegraphics[width=2.5in]{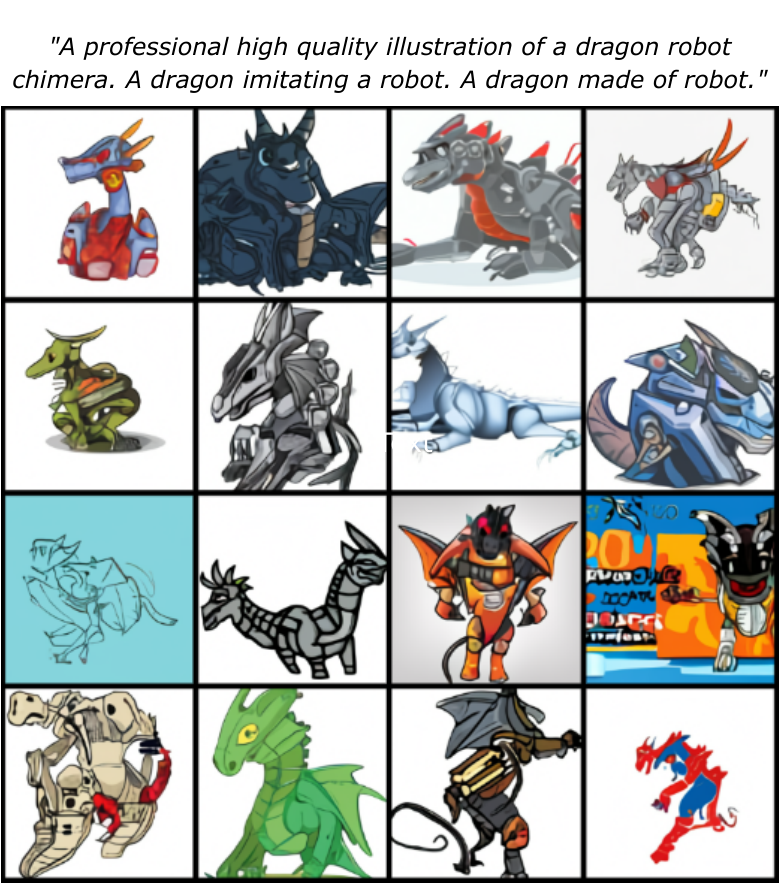}
\caption{Example images produced by DALL-E \cite{ramesh2021}, retrieved from the original OpenAI blog https://openai.com/blog/dall-e/.}
\label{dalle}
\end{figure}

\hfill\break
{\bf{DALL-E.}} Early in 2021, OpenAI released their groundbreaking model \emph{DALL-E} (named after Pixar’s Wall-e and Surrealist painter Salvador Dali) on their blog. Shortly after, they detailed the workings of their model in their paper titled \emph{“Zero-shot Text-to-Image generation”} \cite{ramesh2021}. DALL-E combines a discrete variational autoencoder (dVAE) \cite{rolfe2017}, which learns to map images to lower dimensional tokens, and a Transformer, which autoregressively models text and image tokens (see Fig.~\ref{comparisonTransformers}). 
In this manner, DALL-E is optimized to jointly model text, accompanying images and their token representations. Given a text description as input, DALL-E can predict the image tokens and decode them into an image during inference. Here, zero-shot refers to the ability to generate images based on text descriptions that are not seen during training. This means that DALL-E can combine concepts that it has learned separately but never seen together in a single generated image. For example, it has seen both robots and illustrations of dragons in the training data but it has not seen a robot in the shape of a dragon. However, when prompted to generate “a robot dragon”, the model can produce sensible images (see Fig.~\ref{dalle}). This remarkable capability of the model has resulted in hype surrounding DALL-E.
Although DALL-E can generate cartoons and art styles quite well, it lacks accuracy when generating photo-realistic images. As a result, OpenAI and other companies have devoted substantial resources to create an improved text-to-image model.

{\bf{CogView.}} Concurrently with DALL-E, \cite{ding2021} created \emph{CogView}, a similar text-to-image model that supports Chinese rather than English. Besides the innovative idea to combine VAE and transformers, they include other features such as super-resolution to improve the resolution of the generated images. Despite their super-resolution module, their generated images lack photo-realistic quality.

{\bf{Make-A-Scene.}} In 2022, Meta AI released their \emph{Make-A-Scene} model \cite{gafni2022}. Their Transformer-based text-to-image model allows the user more control over the generated image by working with segmentation maps. During training, the model receives a text prompt, segmentation map and accompanying image as input (similarly as GauGAN2).  The model then learns a latent mapping based on tokenized versions of the inputs. During inference, Make-A-Scene is able to generate an image and segmentation map based solely on text input (see Fig.~\ref{comparisonTransformers}).  Alternatively, one can provide a segmentation map as input to steer the desired output. Moreover, one can alter the segmentation map that the model produces to steer the image generation. 

\begin{figure}[ht]
\centering
\includegraphics[width=3.8in]{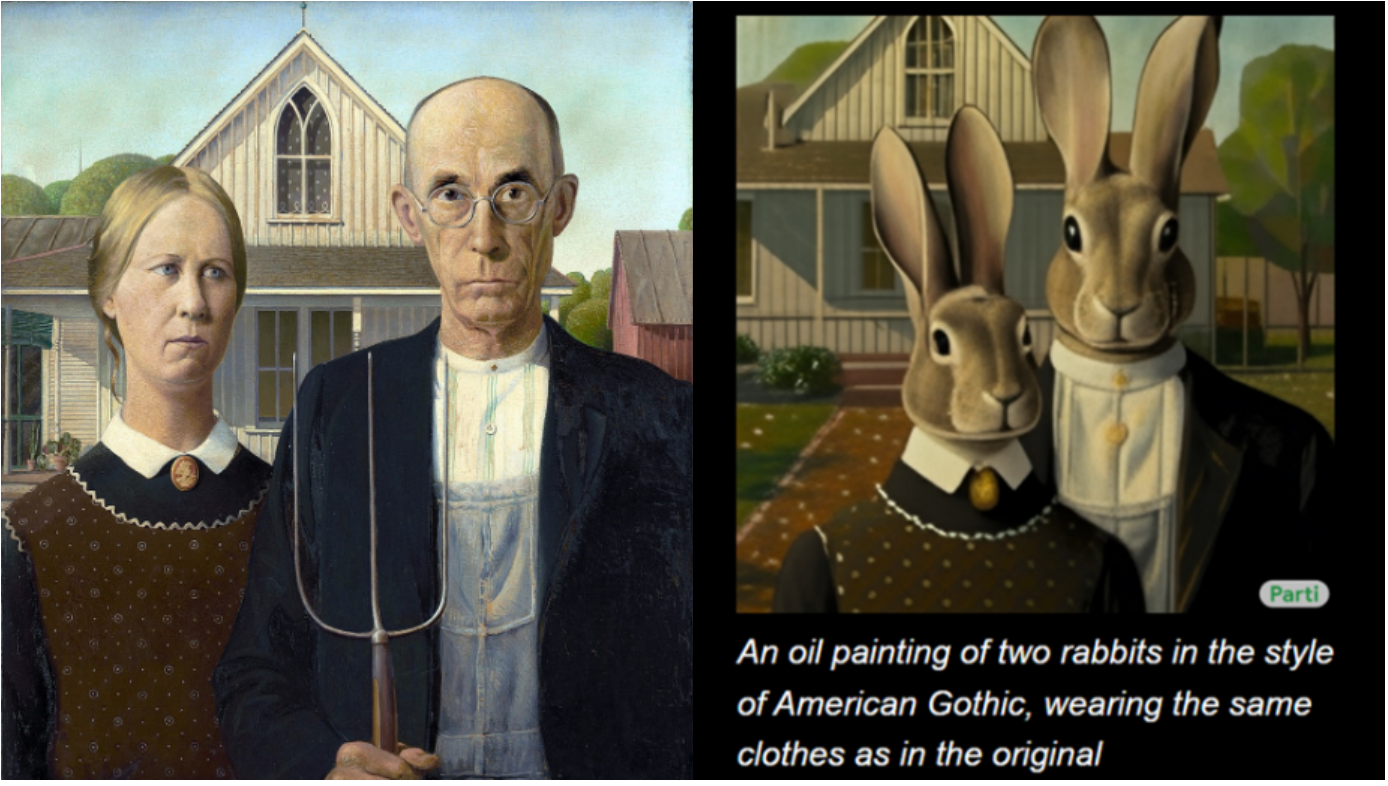}
\caption{\emph{(Left)} \emph{American Gothic} by Grant Wood. \emph{(Right)} An example image generated by Parti \cite{yu2022}, retrieved from https://parti.research.google/.}
\label{parti}
\end{figure}

{\bf{Parti.}} Later in 2022, Google released their Transformer-based text-to-image model called \emph{Parti} which stands for \emph{Pathways Autoregressive Text-to-Image model} \cite{yu2022}. This was the second text-to-image model Google released that year, a month after releasing Imagen \cite{saharia2022} (see Section \ref{diffusion_models}). Parti is based on a ViT-VQGAN \cite{yu2022a}, which combines the transformer encoder and decoder with an adversarial loss of a pretrained GAN to optimize image generation. Parti uses an additional transformer encoder to handle the text input, which is transformed into text tokens to serve as input to a transformer decoder alongside the image tokens from the ViT-VQGAN during training. At test time, the Transformer solely receives text as input and predicts the image tokens, which are then provided to the ViT-VQGAN to detokenize and turn them into an actual image (see Fig.~\ref{comparisonTransformers}).  
Parti outperforms all other text-to-image models on FID score, with a zero-shot FID score of 7.23 and a fine-tuned FID score of 3.22 on MS-COCO \cite{lin2014} (see Table~\ref{table}). Fig.~\ref{parti} is a great example of Parti's remarkable capability to extract the essence of what the caption refers to (in this case, the \emph{style} of American Gothic), rather than simply generating a copy of the original in which the rabbits seem photoshopped. 

{\bf{Muse.}} Early in 2023, Google released \emph{Muse} \cite{chang2023}, another Transformer-based text-to-image model. Muse includes a VQGAN tokenizer to transfer images into tokens and vice versa. The text input is turned into tokens using a pre-trained transformer encoder of the T5-XXL language model \cite{raffel2020} (see Fig.~\ref{comparisonTransformers}). The advantage of using such a pre-trained encoder is that this model component has been trained on a large corpus of text. One struggle in training text-to-image models is the time consuming gathering of a large set of high quality image and caption pairs. Therefore, the text/caption pairs likely only span a limited section of the known concepts within a language. Since NLP models are trained solely with text as input (which is easier to gather), they are trained on a more encompassing corpus of text. The authors find that the inclusion of the pretrained encoder results in higher quality, photo-realistic images and that the generation of text in the generated images is more accurate. Since then, other text-to-image models have incorporated T5-XXL as text encoder in a similar manner (see Section~\ref{diffusion_models} on diffusion models).

Muse achieves an FID score of 7.88 along with a CLIP score of 0.32, which is a measure of how close the generated images are to the prompted caption. Different from the FID score, which measures image realism by comparing the feature distributions of generated and real images, the CLIP score evaluates text–image alignment. It leverages the CLIP model to embed text and images into a shared space and quantifies how well a generated image matches the input prompt, with higher scores indicating stronger semantic consistency. \cite{chang2023} verified this further in a behavioral study in which human raters indicated that the Muse-generated images are better aligned with the prompts compared to images generated with Stable Diffusion (see Section~\ref{diffusion_models}). In addition to image generation, Muse allows inpainting, outpainting and mask-free editing (which will be explained in more detail in Section~\ref{diffusion_models}).

\subsubsection{Diffusion Models}\label{diffusion_models}
\begin{figure*}[!t]
\centering
\includegraphics[width=5.5in]{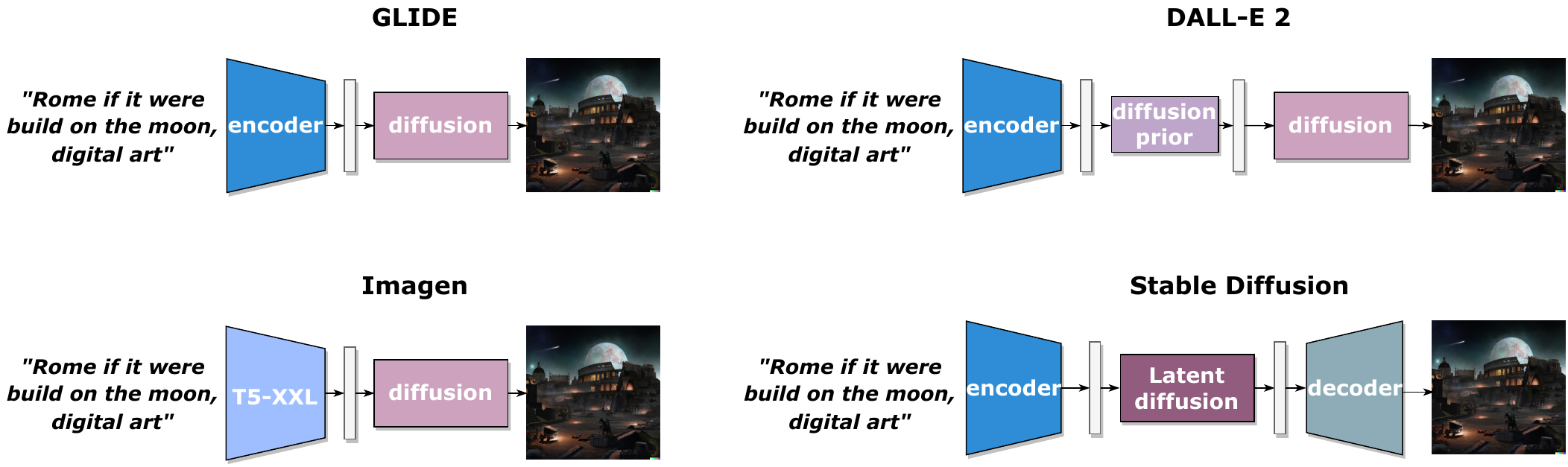}
\caption{Comparison of the diffusion model-based text-to-image models.}
\label{comparisontexttoimage}
\end{figure*}

\hfill\break 
{\bf{GLIDE.}} In 2021, OpenAI came out with a paper that showed that diffusion models outperform GANs on image generation \cite{dhariwal2021}. Less than a year after, OpenAI applied this insight to text-to-image generation, and they released \emph{GLIDE} \cite{nichol2021}, a pipeline consisting of a diffusion model for image synthesis and a transformer encoder for text input (see Fig.~\ref{comparisontexttoimage}). This new and improved model is trained on the same dataset as DALL-E, yet the quality of their generated images strongly outperforms DALL-E. In a study where they asked human participants to judge the generated images from DALL-E and GLIDE, the raters preferred the GLIDE images over the DALL-E images 87\% of the time for photorealism and 69\% of the time for caption similarity. In addition, they preferred blurred GLIDE images over reranked or cherry picked DALL-E images. 

\begin{figure}[ht]
\centering
\includegraphics[width=3.8in]{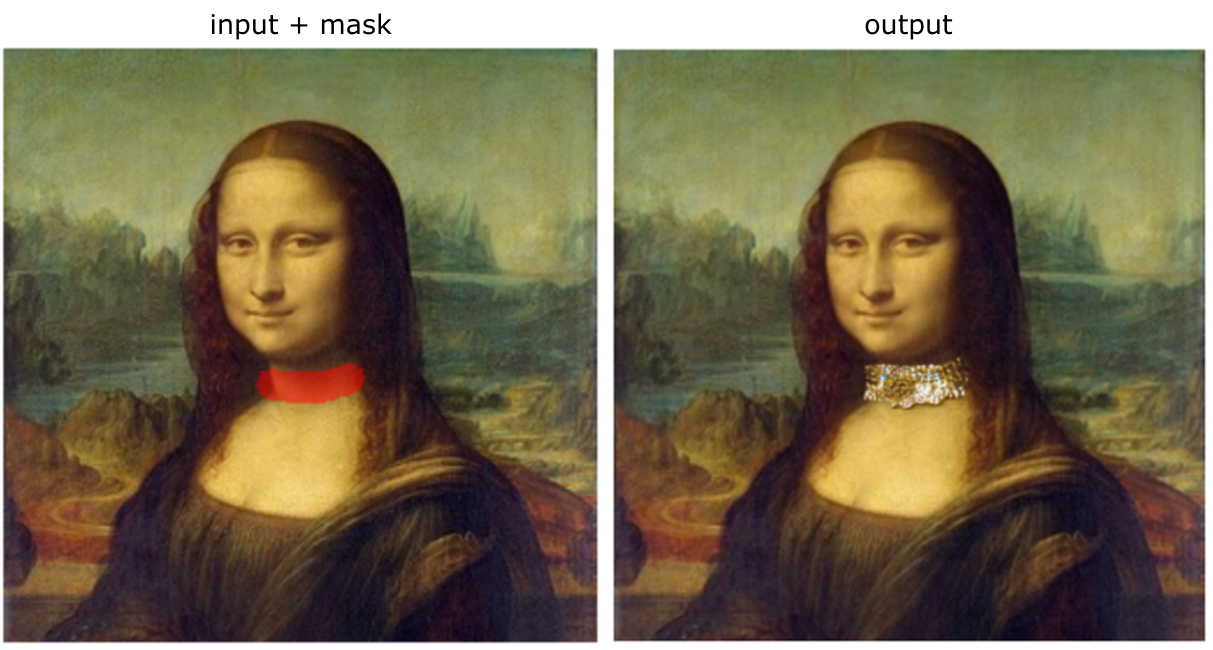}
\caption{An example of inpainting with GLIDE. It receives an image of the \emph{Mona Lisa} as input as well as a mask and a text description ``a golden necklace''. Then, it generates the output as \emph{Mona Lisa} with a golden necklace. \cite{nichol2021}.}
\label{glide}
\end{figure}

Besides improving photorealism, GLIDE also offers the feature of image inpainting, meaning that you can edit a specific region in an existing or computer-generated image. For example, one can take an image of the \emph{Mona Lisa} and add a necklace by providing the image and a text description (e.g., “a golden necklace'') as input to GLIDE (see Fig.~\ref{glide}).

\begin{figure}[ht]
\centering
\includegraphics[width=3.8in]{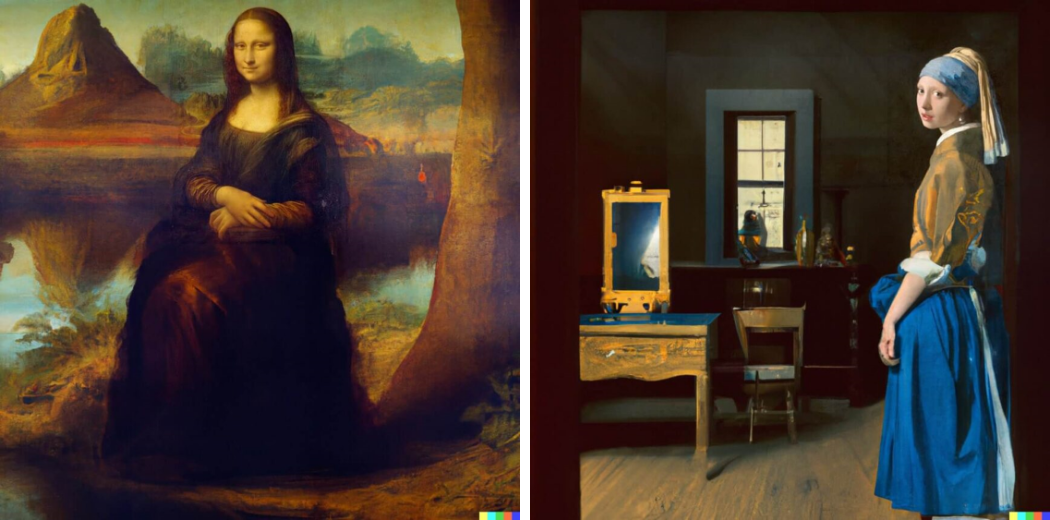}
\caption{Image outpainting examples by DALL-E 2 \cite{ramesh2022}. \emph{(Left)} \emph{Mona Lisa}. \emph{(Right)} \emph{Girl with a Pearl Earring}.}
\label{dalle2}
\end{figure}

{\bf{DALL-E 2.}} Even though GLIDE was an impressive improvement upon DALL-E, it did not garner the same attention. When OpenAI released \emph{DALL-E 2} \cite{ramesh2022} in 2022, an advancement of GLIDE, this changed. DALL-E 2 has a similar diffusion pipeline as GLIDE, but the text input of this diffusion pipeline has been improved. Whereas GLIDE uses an untrained transformer encoder to format the text, DALL-E 2 uses the CLIP text encoder. Additionally, it uses a prior model to transform the text embedding into a CLIP image embedding before feeding it to the diffusion model (see Fig.~\ref{comparisontexttoimage}). This is achieved by transforming images and text descriptions to image and text embeddings (or tokens) respectively using transformer encoders. The model is then trained to link the correct embeddings with one another. The relationship between text descriptions and images is exploited in DALL-E 2, providing the diffusion model with an embedding that reflects the text input while being better suited for image generation. In addition to improving image quality compared to GLIDE, DALL-E 2 allows the user to extend the background of an existing image or computer-generated one (referred to as outpainting, see Fig.~\ref{dalle2}) and to generate variations of images.

{\bf{DALL-E 3.}} An important limitation of current text-to-image models is their difficulty in accurately reflecting everything in the text prompt they receive. A common example to illustrate their lacking prompt following ability is the generation of a penguin cartoon, wearing clothing items in different colors. The models tend to have difficulty linking the right color to the right clothing item. Researchers at OpenAI hypothesized that this limitation stems from a lack of correspondence between text descriptions and images in the datasets used to train these text-to-image models. They trained an image captioner to re-label these datasets with more accurate yet synthetic prompts and found that this approach improved the current text-to-image models. \emph{DALL-E 3} \cite{betker2023} incorporates this re-captioning, amongst other model improvements (such as replacing the text encoder with T5-XXL \cite{raffel2020}, similarly as Muse \cite{chang2023}). At inference time, the model generates an image and a synthetic, rich prompt. The user can then edit the prompt to have more control over the output image. DALL-E 3 is currently available through ChatGPT and Bing image creator.

{\bf{Imagen.}} Shortly after DALL-E 2, Google released their first text-to-image model called \emph{Imagen} \cite{saharia2022}. Their model is closer to GLIDE in architecture, since it does not rely on CLIP embeddings. Instead, it uses the pretrained encoder of the NLP model T5-XXL (similarly as their earlier model Muse), whose embeddings are fed to the diffusion model (see Fig.~\ref{comparisontexttoimage}). As a result, the model is able to generate images that contain text more accurately (which is something OpenAI’s models struggled with). In addition, Imagen feeds the generated image of the diffusion model to two super-resolution diffusion models to increase the resolution of the images.

{\bf{Stable Diffusion.}} The biggest revolution in the field is perhaps the fully open source release of \emph{Stable Diffusion} \cite{rombach2022} by a company called Stability AI (we elaborate on the concerns regarding open source models in Section~\ref{ethical_concerns}). Their main contribution is the computational efficiency of their model, as opposed to the above mentioned text-to-image models. Rather than operating in pixel-space, Stable Diffusion operates in (lower-dimensional) latent space and maps the output of the diffusion process back to pixel space using a VAE (see Fig.~\ref{comparisontexttoimage}). Whereas previous text-to-image models require hundreds of GPU computing days, this latent diffusion model requires significantly smaller computational demands and is therefore more accessible to those with less resources. Besides image generation, Stable Diffusion additionally allows the user to modify existing images through image-to-image translation (e.g., turning a sketch into digital art) or inpainting (removing or adding elements in an existing image). 

{\bf{SDXL.}} Recently, Stability AI released a more powerful and larger version of Stable Diffusion, called \emph{SDXL} \cite{podell2023}. This model still operates in latent space but has significantly more parameters than the earlier Stable Diffusion releases. This is due to the use of more powerful pre-trained text encoders as well as an increase in transformer blocks in the U-Net part of the diffusion model. In addition, SDXL features a new component called the ‘refiner’, which is a U-Net placed in between the output of the latent diffusion and the VAE to map the latent representation to pixelspace. By placing a refiner in between these two steps, the model learns to further refine the image latent before it is mapped back to pixel space. They illustrate that this addition results in images with more fine-grained details, especially in the background. Besides these changes to the diffusion models architecture, they also add new features that the model is conditioned on. This conditioning allows more control over the generated images. For example, SDXL is conditioned on image size and cropping parameters, allowing the user to specify the desired resolution of the output image and how centered the central object should be. As such, SDXL can generate images with a different height and width, which is common in artworks and photographs.

To mitigate the limitation that diffusion models perform multiple steps at inference time to generate an image from noise, various distillation methods have been proposed to generate images with few steps. These distillation methods operated in pixel-space, as most initial diffusion models do. Stability AI proposed a latent distillation method to go with their latent diffusion models that is more computationally efficient \cite{sauer2024}. They coined these models \emph{SD Turbo} and \emph{SDXL Turbo}. Their Turbo models are on par or outperform other leading text-to-image models (without distillation method) on image quality and prompt alignment as assessed with a user study.

{\bf{FLUX.}} It is a text-to-image model developed by Black Forest Labs, a start-up based in Germany that was founded by researchers previously involved in the development of various Stable Diffusion versions. FLUX is best known as the image generation engine behind Grok, X's (formerly Twitter) built-in image generator. The model is a rectified flow transformer, which operates similarly to a diffusion model but with greater efficiency \cite{esser2024scaling}. Traditional diffusion models typically require many iterative steps to transform noise into a data sample. In contrast, rectified flow models reduce the number of required steps by learning a straight line between noise to data. Traditional diffusion models learn by gradually adding Gaussian noise to an image and then learning to denoise it. The Gaussian noise is stochastic, resulting in a curved path between noise and data that requires many steps. On the other hand, rectified flow transformers construct a synthetic path between noise and data. Their learning objective is to transform random noise into data by traveling along these straight line trajectories and learning a velocity vector. Each step involves a linear interpolation between noise and data, enabling faster inference with fewer steps. FLUX offers inpainting, outpainting, and image variation in addition to text-to-image generation. It outperforms previous models in quantitative analyses and human preference ratings.

\begin{figure}[ht]
\centering
\includegraphics[width=2in]{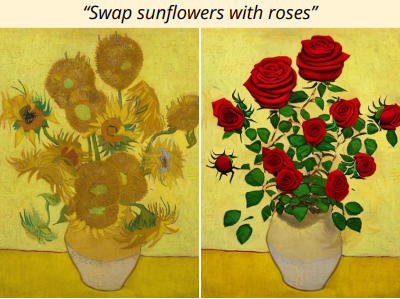}
\caption{InstructPix2Pix turns the \emph{Sunflowers} by Vincent Van Gogh into a painting of roses \cite{brooks2022}.}
\label{sunflower}
\end{figure}

{\bf{InstructPix2Pix.}}  Although several text-to-image models have the inpainting feature, in practice it is difficult to get the desired output based on a text description. The user either needs to describe the entire output image, or create a mask for the area in the image that should be modified and describe the desired modification. \emph{InstructPix2Pix} \cite{brooks2022} is a modification of Stable Diffusion and allows the user to modify images through intuitive text instructions. Rather than having to describe the desired output image or providing a mask, the user can just write an intuitive instruction of how the input image should be adjusted (mask-free editing). For example, if you would like to turn the \emph{Sunflowers} by Vincent Van Gogh into a painting of roses, you can just write the instruction “Swap sunflowers with roses” (Fig.~\ref{sunflower}).

\section{Comparison of deep generative models}\label{sec4}
Due to the wide variety of models that have been recently developed in the field of generative AI, it may be hard to decide which model is most suited for one's purposes. Table~\ref{table} provides a comparison of the above mentioned models in terms of their size (trainable parameters), the quality of the generated images (FID score), the capabilities of the model and its accessibility (open source vs. not for public use). It should be noted that FID score \cite{heusel2017} is a suboptimal measure of image quality and especially image aesthetics. In fact, \cite{kirstain2023} found that FID score correlates negatively with human preferences for prompt-based generated images. They propose an alternative measure, called \emph{PickScore}, that correlates highly with human preferences. We should also note that large diffusion models require substantial computational resources and energy, raising sustainability concern \cite{bertazzini2025hidden}. 

\begin{table*}[ht]
\centering
\caption{Summary of deep generative models}\label{table}
\begin{threeparttable}
\begin{tabular}{@{}lllll@{}}
\toprule
Model & \thead[l]{FID on\\ MS-COCO}  & \thead[l]{Trainable\\ parameters} & \thead[l]{Open\\ source} & Capabilities \\
\midrule
ArtGAN \cite{tan2017} & -\tnote{*} & -   & Yes & \makecell[l]{Image generation} \\
CAN  \cite{elgammal2017}   & - & -  & Yes & \makecell[l]{Image generation}  \\
pix2pix  \cite{isola2017} & - & -  & Yes & \makecell[l]{Image manipulation}  \\
CycleGAN  \cite{zhu2017} & - & -  & Yes & \makecell[l]{Image manipulation} \\
GauGAN  \cite{park2019} & 22.6 & -  & Yes & \makecell[l]{Image generation,\\ image manipulation,\\ text-to-image} \\
LAFITE  \cite{zhou2021} & 26.94 & 75M  & Yes & \makecell[l]{Text-to-image} \\
DALL-E \cite{ramesh2021}  & 27.50 & 12B  & No & \makecell[l]{Text-to-image} \\
CogView  \cite{ding2021} & 27.10 & 4B  & Yes & \makecell[l]{Text-to-image} \\
Make-A-Scene \cite{gafni2022} & 11.84 & 4B &  No & \makecell[l]{Text-to-image,\\ image manipulation} \\
Parti  \cite{yu2022} & 7.23 & 20B  & No & \makecell[l]{Text-to-image}\\
Muse  \cite{chang2023}  & 7.88 & 3B  & No & \makecell[l]{Text-to-image, inpainting,\\  outpainting, mask-free editing} \\
GLIDE  \cite{nichol2021}  & 12.24 & 3.5B  & Partially & \makecell[l]{Text-to-image, inpainting} \\
DALL-E 2  \cite{ramesh2022}  & 10.39 & 4.5B &  Partially & \makecell[l]{Text-to-image, inpainting,\\ outpainting, generate variations} \\
DALL-E 3  \cite{betker2023} & - & - &  Partially & \makecell[l]{Text-to-image, inpainting,\\ outpainting, generate variations} \\
Imagen  \cite{saharia2022} & 7.27 & 2B  & No & \makecell[l]{Text-to-image}   \\ 
Stable Diffusion  \cite{rombach2022} & 12.63 & 1.45B  & Yes & \makecell[l]{Text-to-image, inpainting,\\ image manipulation}  \\
SDXL  \cite{podell2023} & Various & 2.6B  & Yes & \makecell[l]{Text-to-image, inpainting,\\ image manipulation}  \\
FLUX  \cite{esser2024scaling} & - & 12B  & Yes & \makecell[l]{Text-to-image, inpainting,\\ outpainting, generate variations}  \\
InstructPix2Pix  \cite{brooks2022} & - & -  & Yes & \makecell[l]{Mask-free editing}    \\ 
\bottomrule
\end{tabular}
\begin{tablenotes}
\footnotesize
\item[*] Not reported.
\end{tablenotes}
\end{threeparttable}
\end{table*}

\begin{figure*}[!t]
    \centering

    \begin{minipage}{0.27\textwidth}
        \centering
        \textbf{DALL-E 3}
    \end{minipage}
    \hfill
    \begin{minipage}{0.27\textwidth}
        \centering
        \textbf{SDXL}
    \end{minipage}
    \hfill
    \begin{minipage}{0.27\textwidth}
        \centering
        \textbf{Midjourney}
    \end{minipage}
    
    \vspace{0.1cm} 

    \begin{minipage}{0.27\textwidth}
        \centering
        \includegraphics[width=\textwidth]{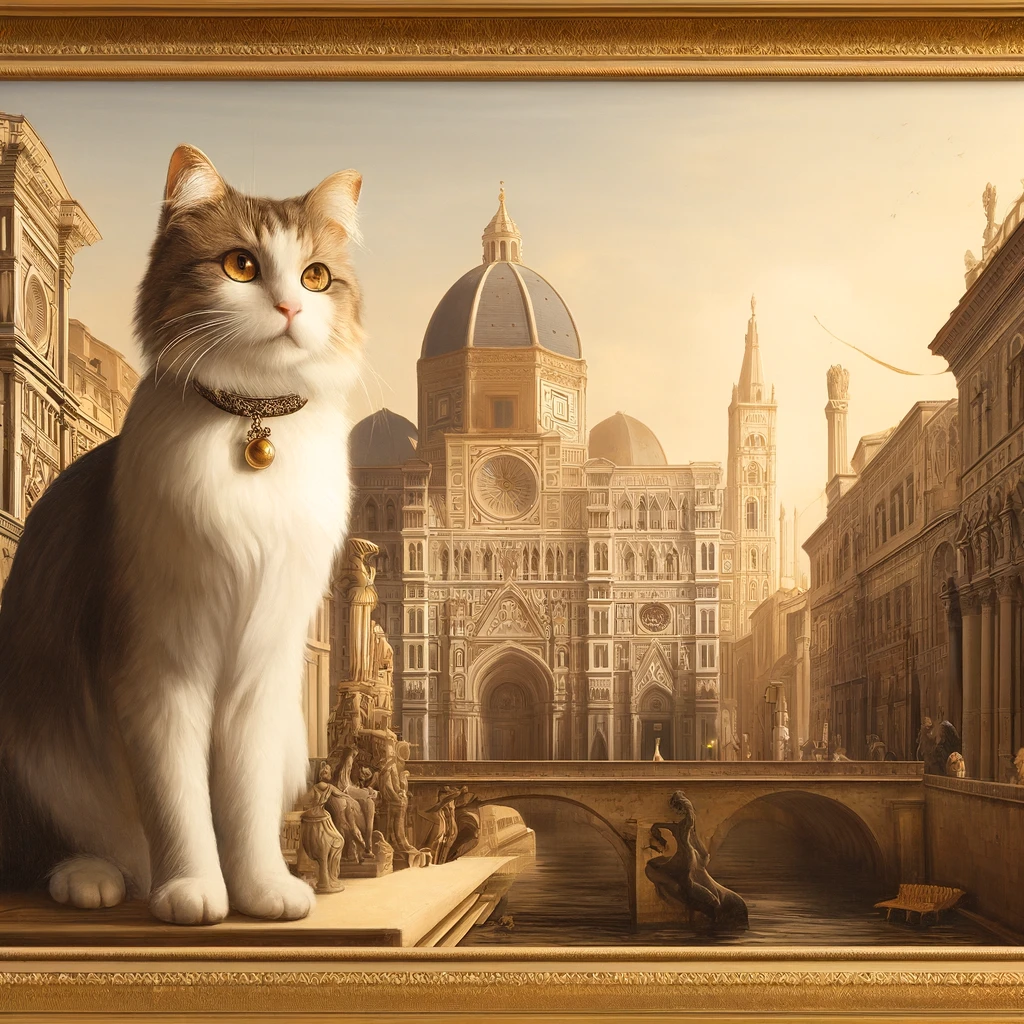}
    \end{minipage}
    \hfill
    \begin{minipage}{0.27\textwidth}
        \centering
        \includegraphics[width=\textwidth]{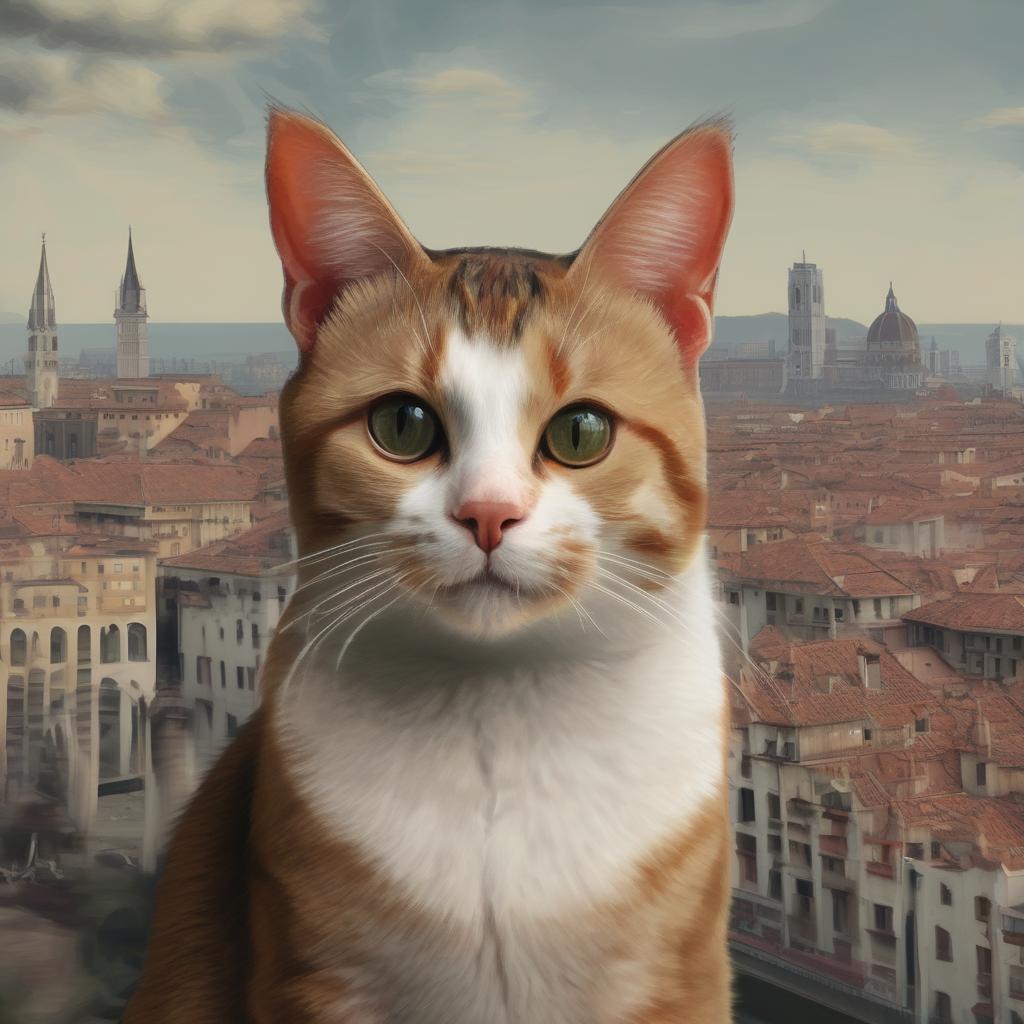}
    \end{minipage}
    \hfill
    \begin{minipage}{0.27\textwidth}
        \centering
        \includegraphics[width=\textwidth]{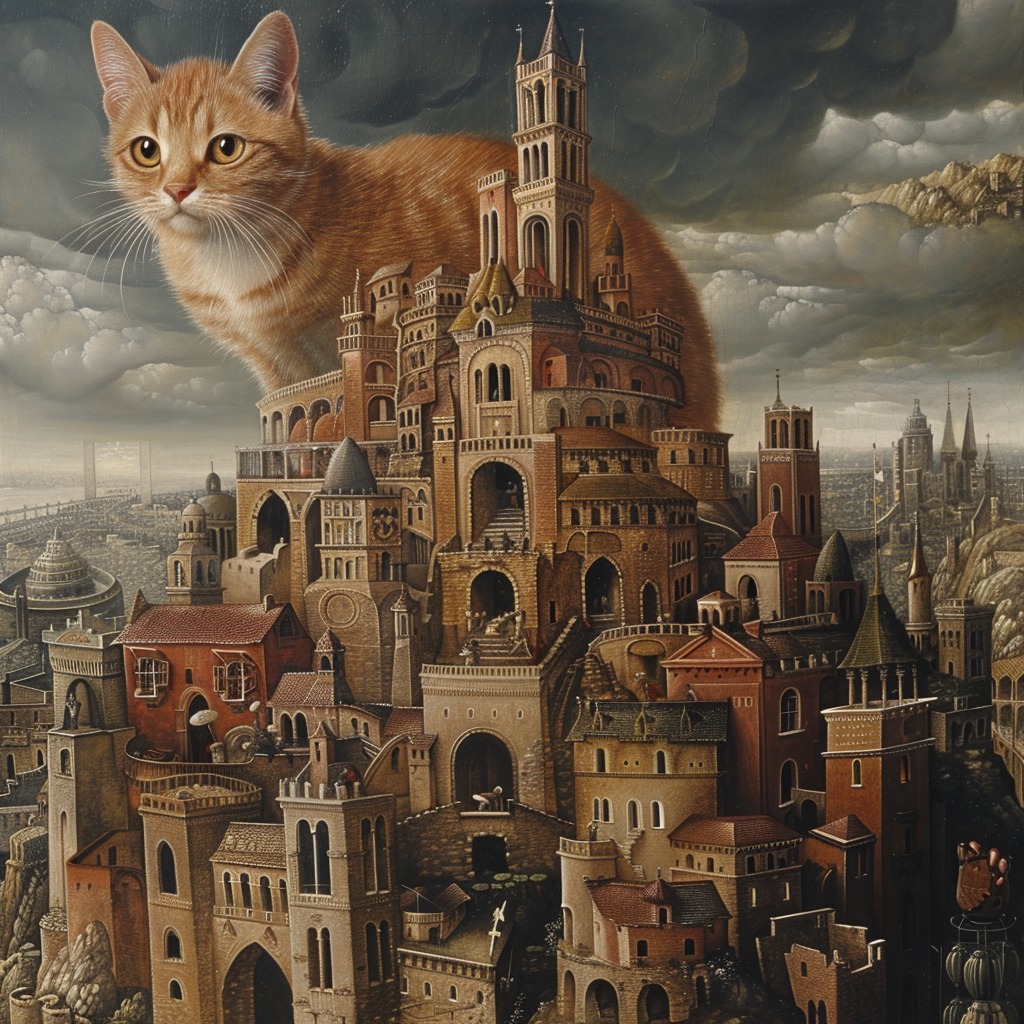}
    \end{minipage}
    \vspace{0.0cm}
    
    \par\smallskip
    \emph{``A Renaissance painting depicting a cat in a cityscape."}
    \par\smallskip
    
    \begin{minipage}{0.27\textwidth}
        \centering
        \includegraphics[width=\textwidth]{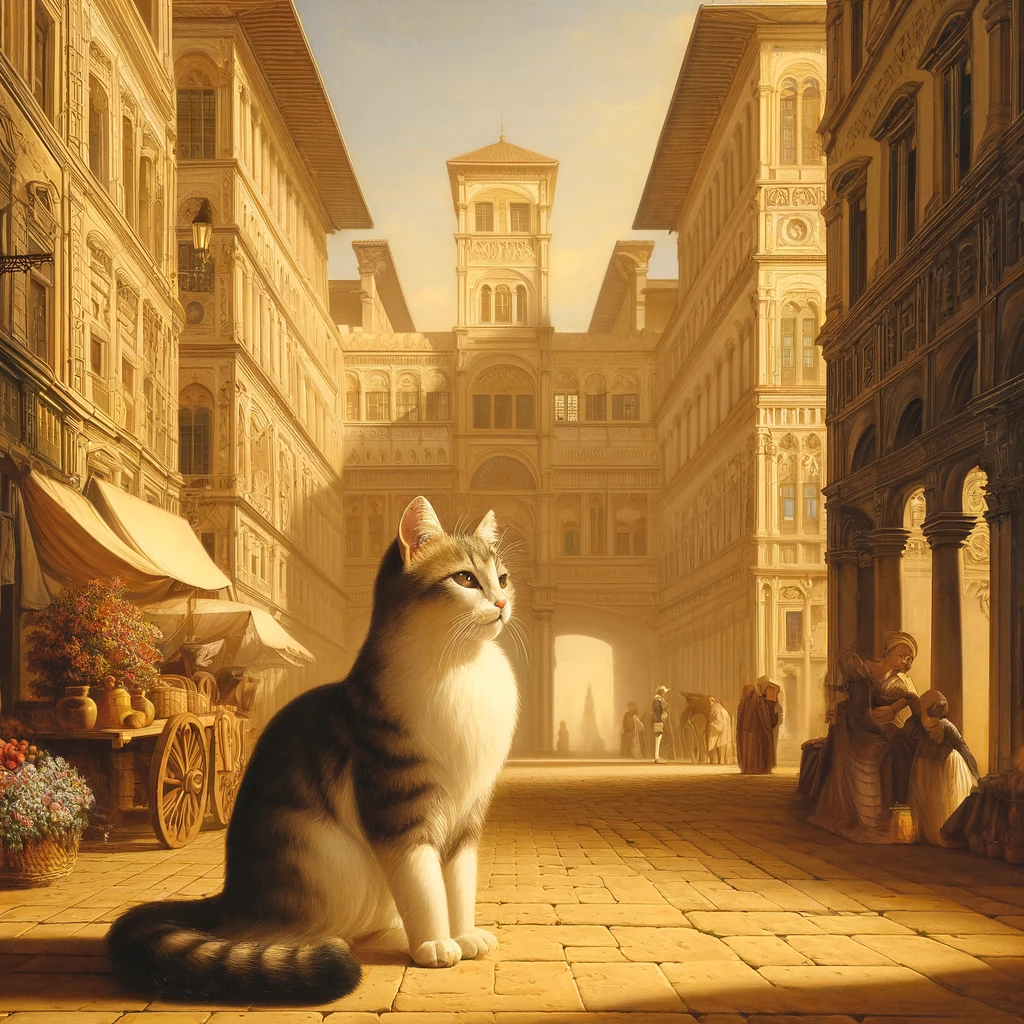}
    \end{minipage}
    \hfill
    \begin{minipage}{0.27\textwidth}
        \centering
        \includegraphics[width=\textwidth]{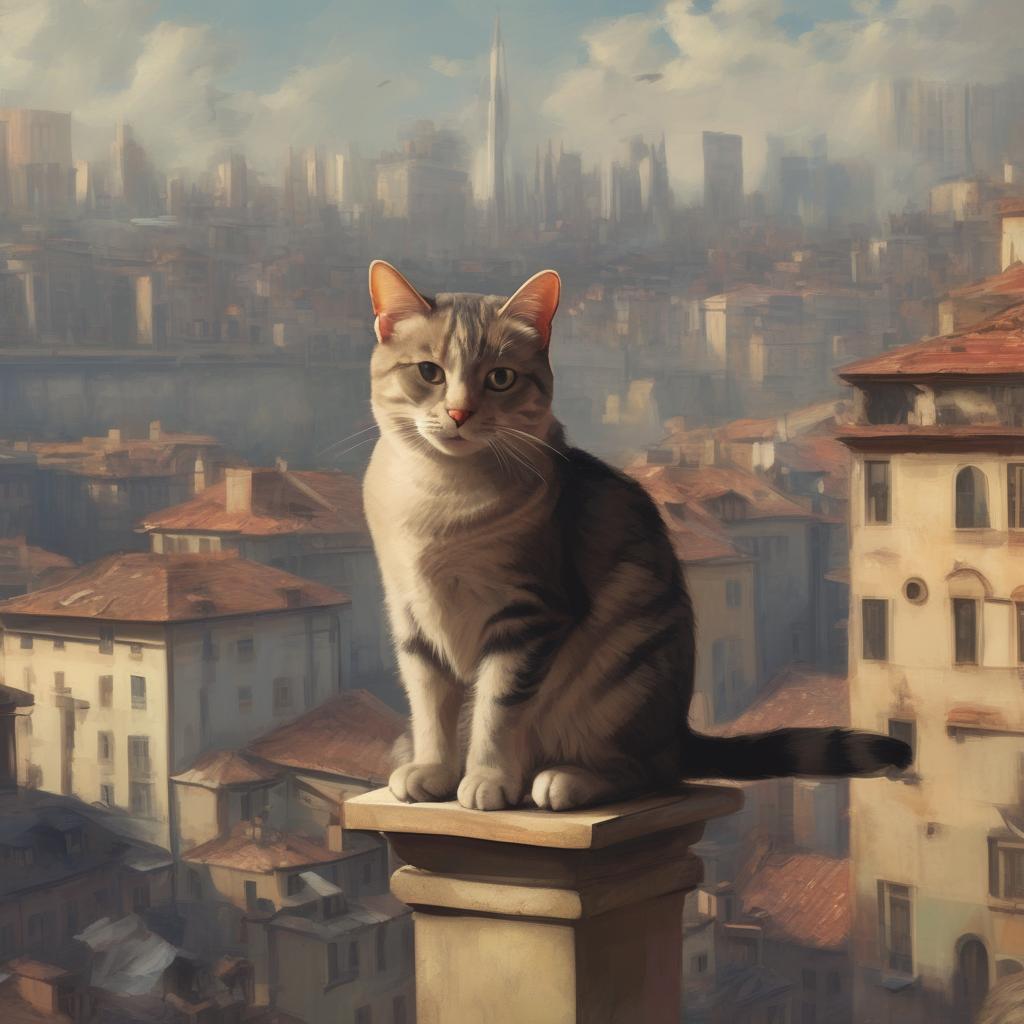}
    \end{minipage}
    \hfill
    \begin{minipage}{0.27\textwidth}
        \centering
        \includegraphics[width=\textwidth]{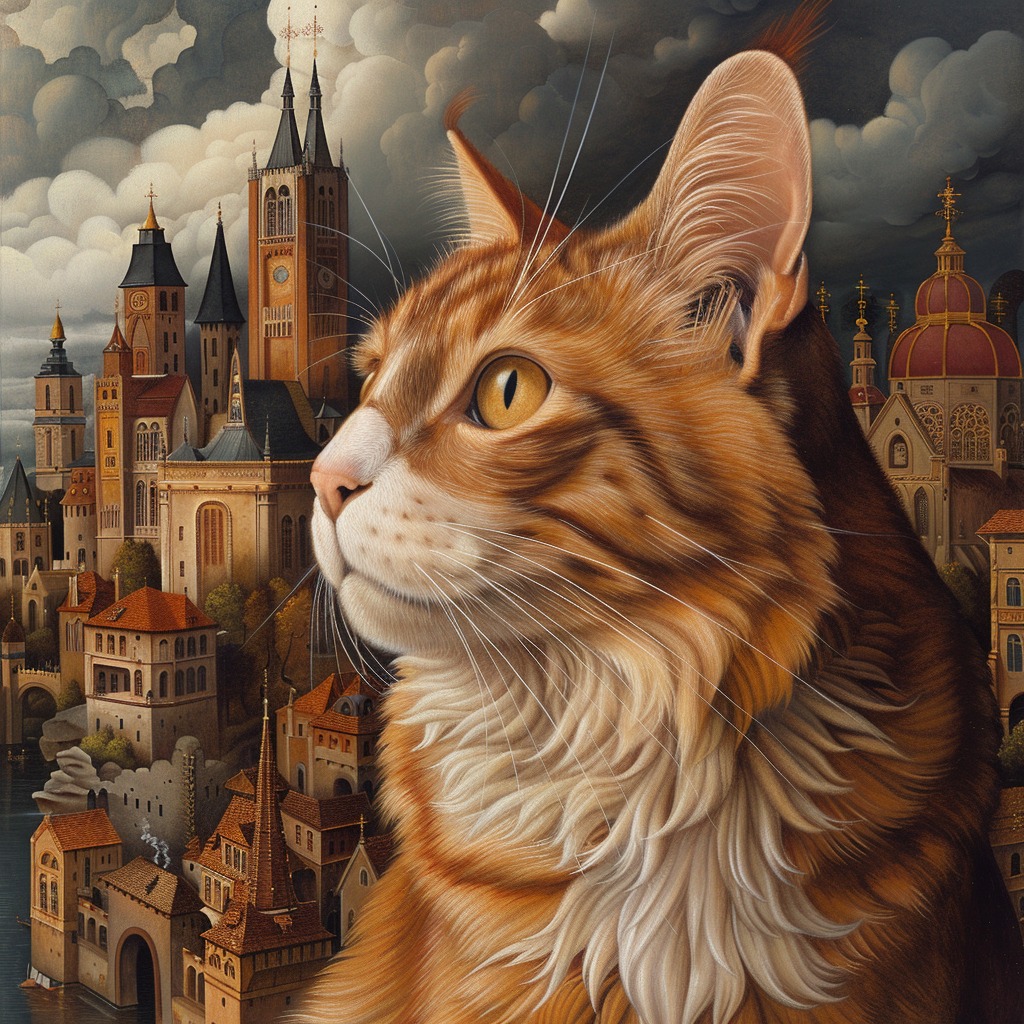}
    \end{minipage}
    \vspace{0.0cm}
    
    \par\smallskip
    \emph{``A highly aesthetic Renaissance painting depicting a cat in a cityscape."}
    \par\smallskip
    
    \begin{minipage}{0.27\textwidth}
        \centering
        \includegraphics[width=\textwidth]{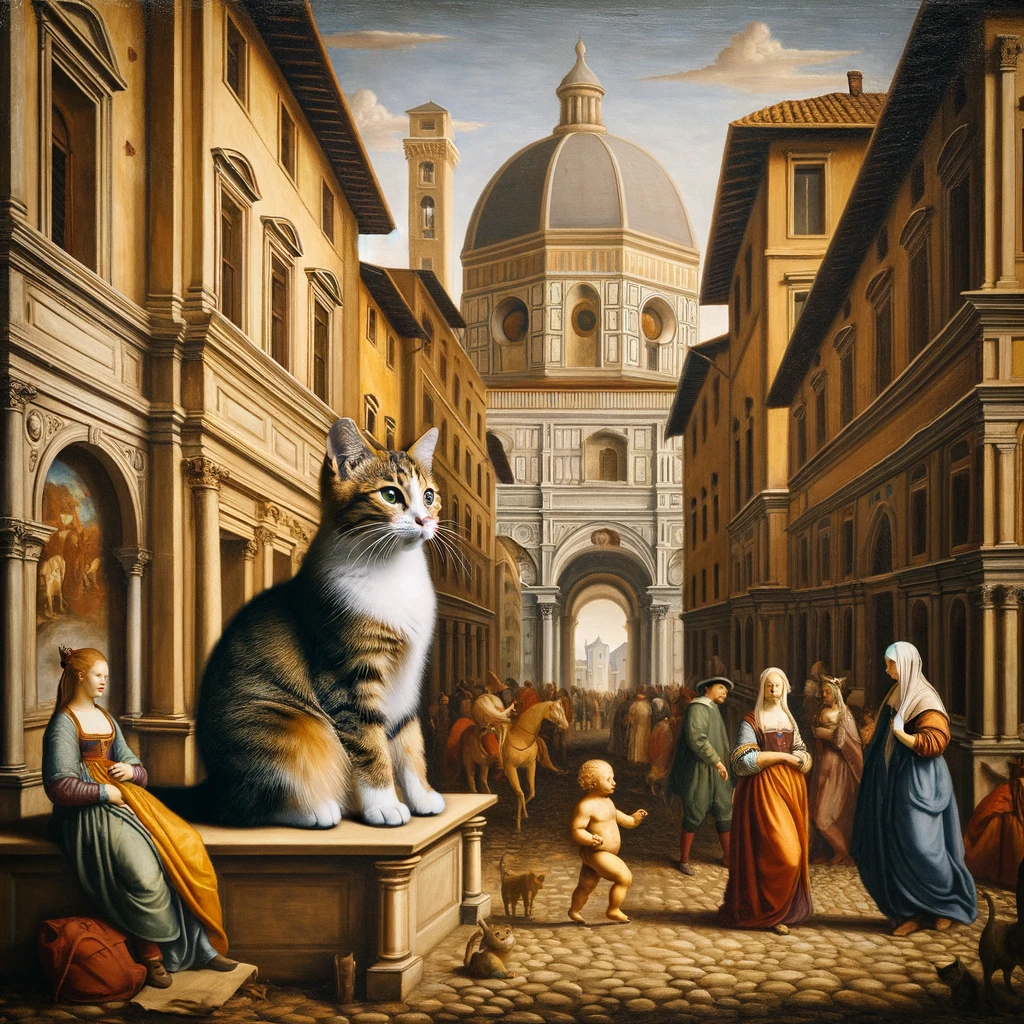}
    \end{minipage}
    \hfill
    \begin{minipage}{0.27\textwidth}
        \centering
        \includegraphics[width=\textwidth]{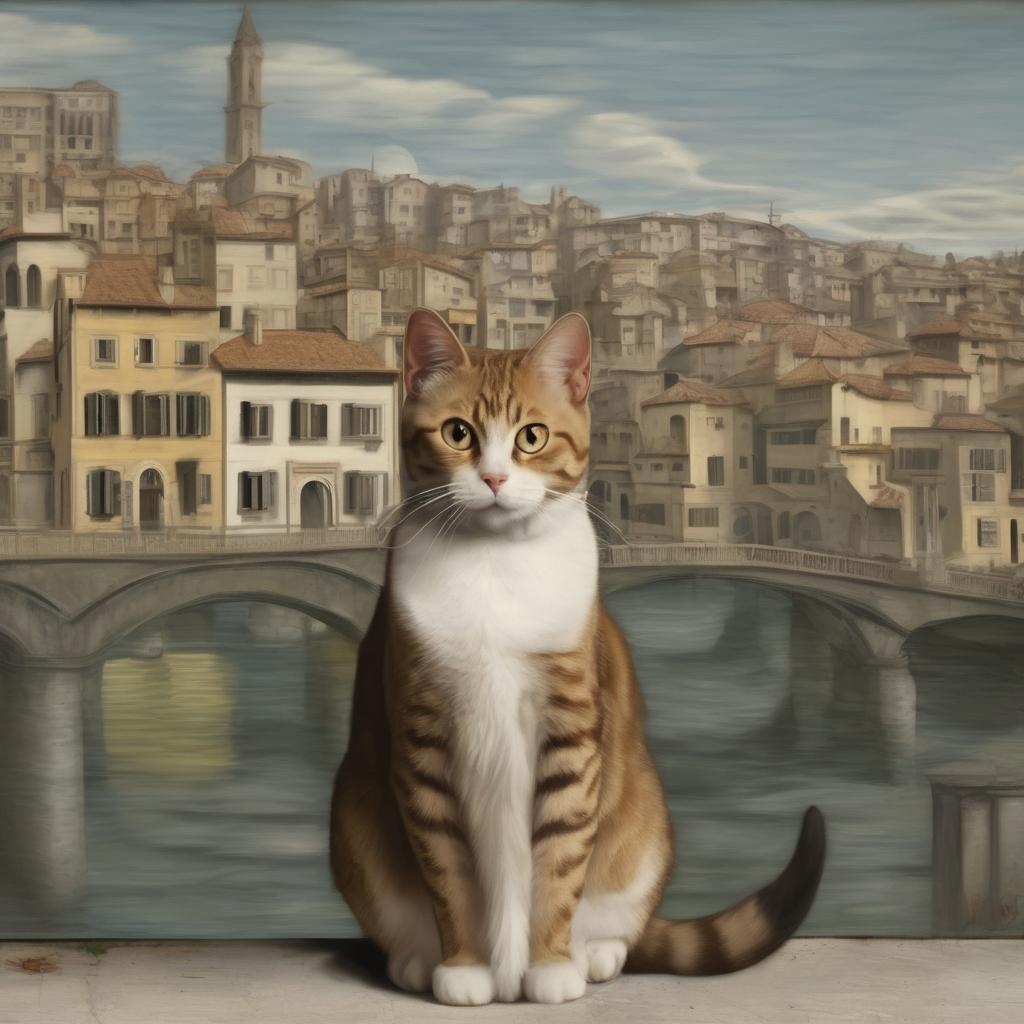}
    \end{minipage}
    \hfill
    \begin{minipage}{0.27\textwidth}
        \centering
        \includegraphics[width=\textwidth]{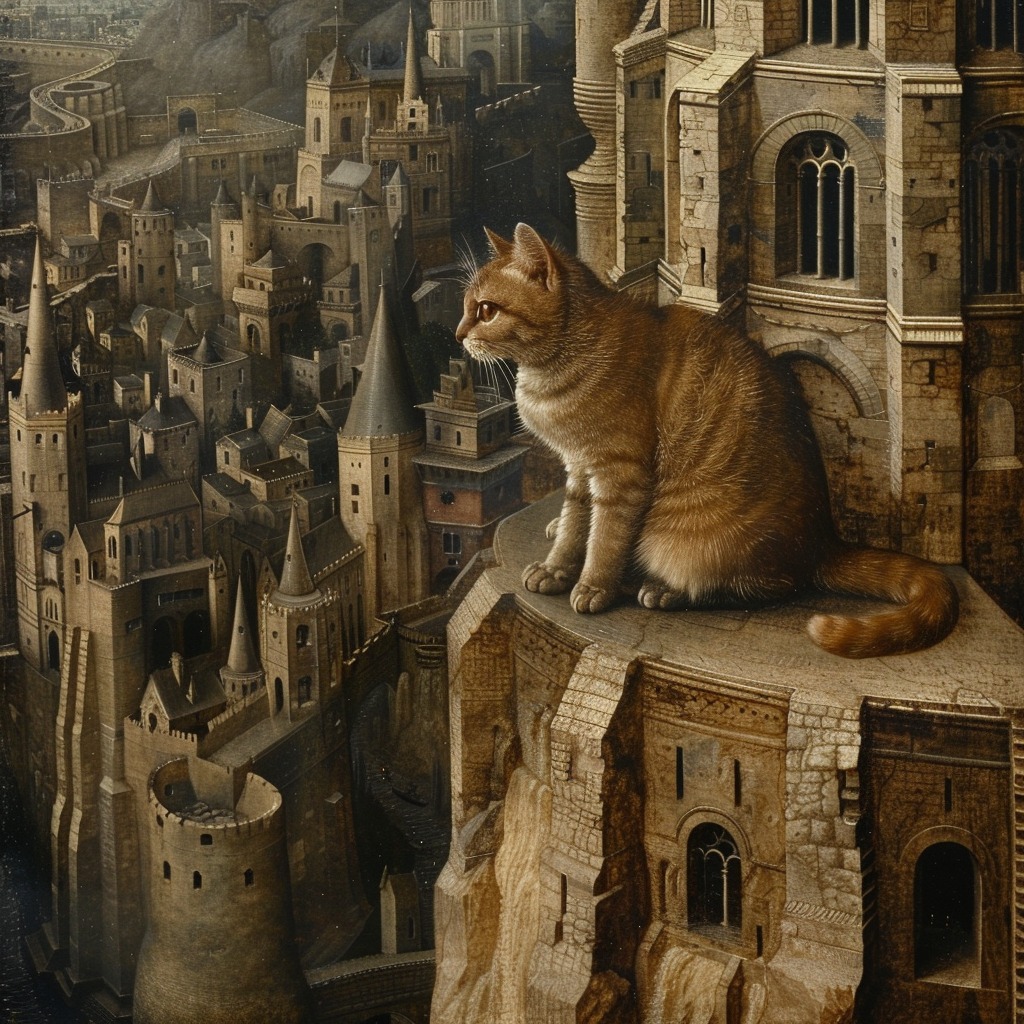}
    \end{minipage}
    \vspace{0.0cm}
    
    \par\smallskip
    \emph{``A Renaissance painting which would be criticized by art critics, depicting a cat in a cityscape."}
    \par\smallskip

    \caption{Images generated by DALL-E 3, SDXL, and Midjourney using the same prompt. \emph{(Top row)} A cat in a detailed Renaissance cityscape with historical architecture. \emph{(Middle row)} Highly aesthetic versions of the top row. \emph{(Bottom row)} Less aesthetic versions of the top row, potentially criticized by art critics.}
    \label{art_comparison}
\end{figure*}

To compare the models further on their capability to generate art-like images, we asked the three widely used models to generate images that represent Renaissance paintings. In addition to DALL-E 3 \cite{betker2023} and SDXL \cite{podell2023}, we include Midjourney in our comparison due to its popularity in text-to-image generation, despite its lack of publicly shared neural network architectures and code. For our comparison, we use prompts to generate images depicting a cat in a cityscape as a Renaissance painting: one highly aesthetic version and another that could potentially be criticized by art critics. Our aim is to see how well these models capture art styles in relation to aesthetics and art criticism. Fig.~\ref{art_comparison} shows the results. All three models seem to capture the Renaissance style for the cityscape aspect of the image but seem to have difficulty applying this style to the cat. This could be explained by cityscape being a common art genre also present in Renaissance paintings. On the other hand, cats are less common in cityscape or Renaissance paintings but are highly common in natural images as part of these models' training dataset. This could explain why the cats are often generated in a more modern drawing style and are disproportionately large compared to the buildings in the images. When we compare the highly aesthetic variants to the neutral prompt images, we can see that the highly aesthetic ones are often enhanced in color, brightness or saturation. \cite{music2023} found that these image features are positively related to human preferences for images generated with a model that aims to enhance the aesthetic value of an image. In a similar line of research, the importance of generating highly aesthetic images has been emphasized by \cite{dai2023}, who propose quality-tuning a latent diffusion model \cite{rombach2022} for this purpose. In Fig.~\ref{art_comparison}, the images for the art criticism prompts cannot be distinguished easily from the neutral prompts, indicating that the models have a weak notion of art criticism, or that they (most probably correctly) assume that any AI-generated image would be criticized by art critics. 

\begin{figure*}[!htbp]
    \centering            
    
    \begin{tabular}{p{0.3\linewidth} p{0.3\linewidth}}
        \centering\small \textbf{SDXL} & \centering\small \textbf{Midjourney} \\
    \end{tabular}
    \vspace{-0.4cm}
    
    \begin{tabular}{cc}
        \includegraphics[width=0.27\linewidth]{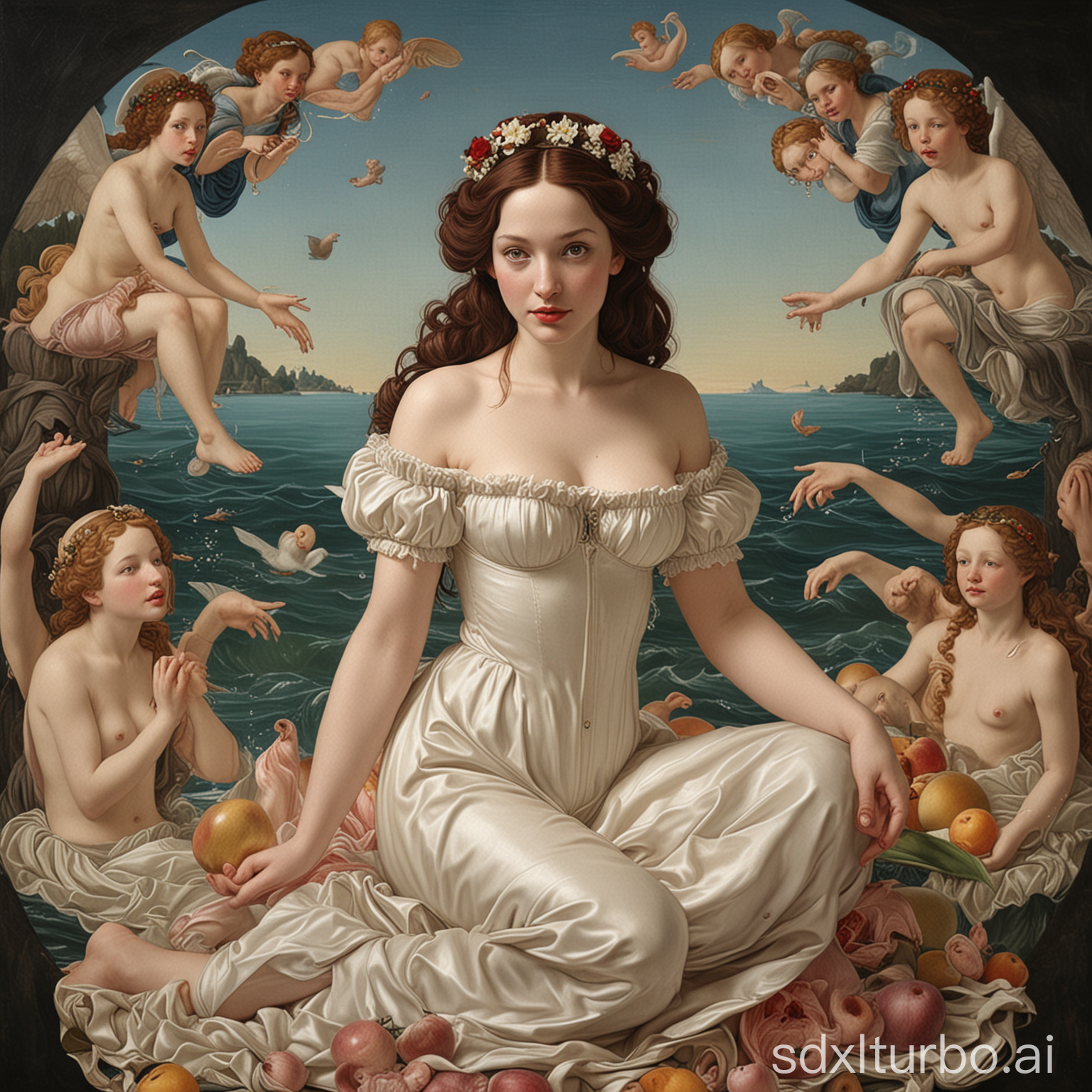} & \includegraphics[width=0.27\linewidth]{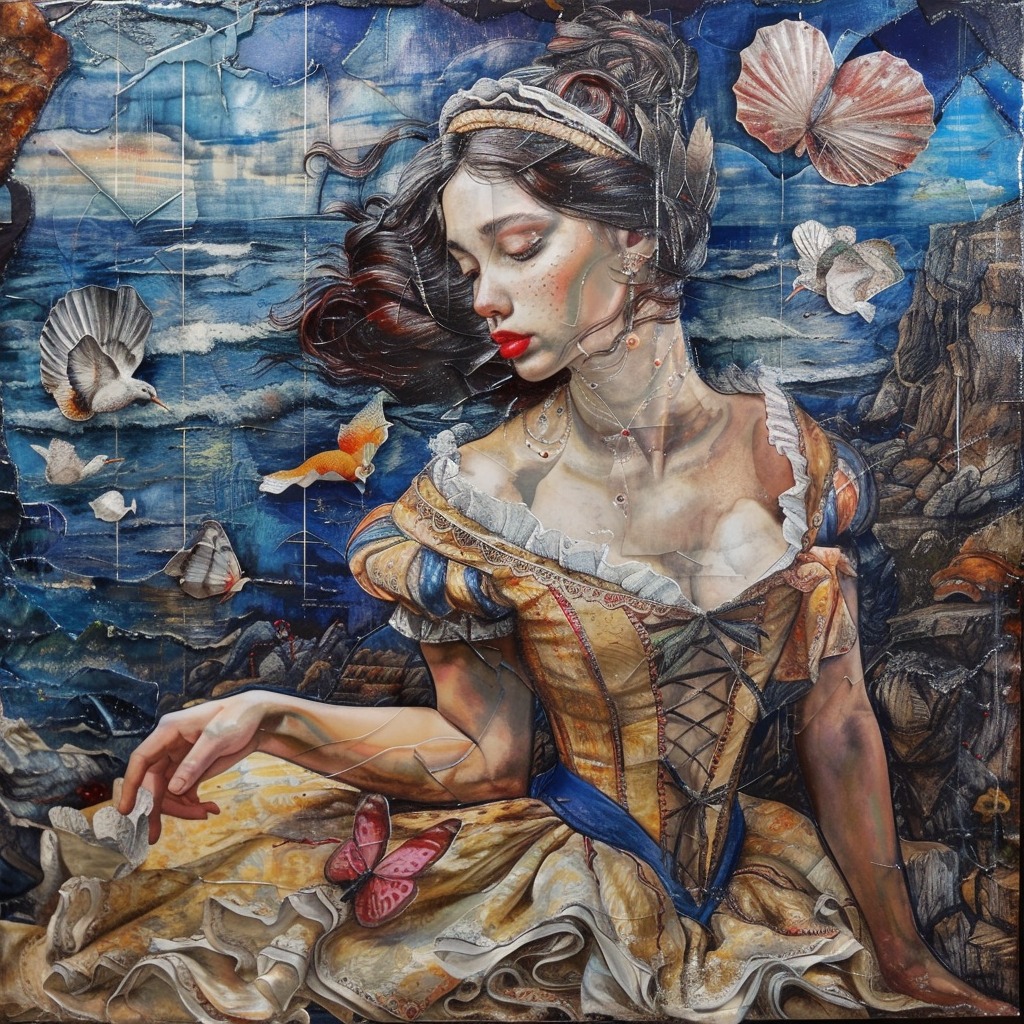} \\
        \includegraphics[width=0.27\linewidth]{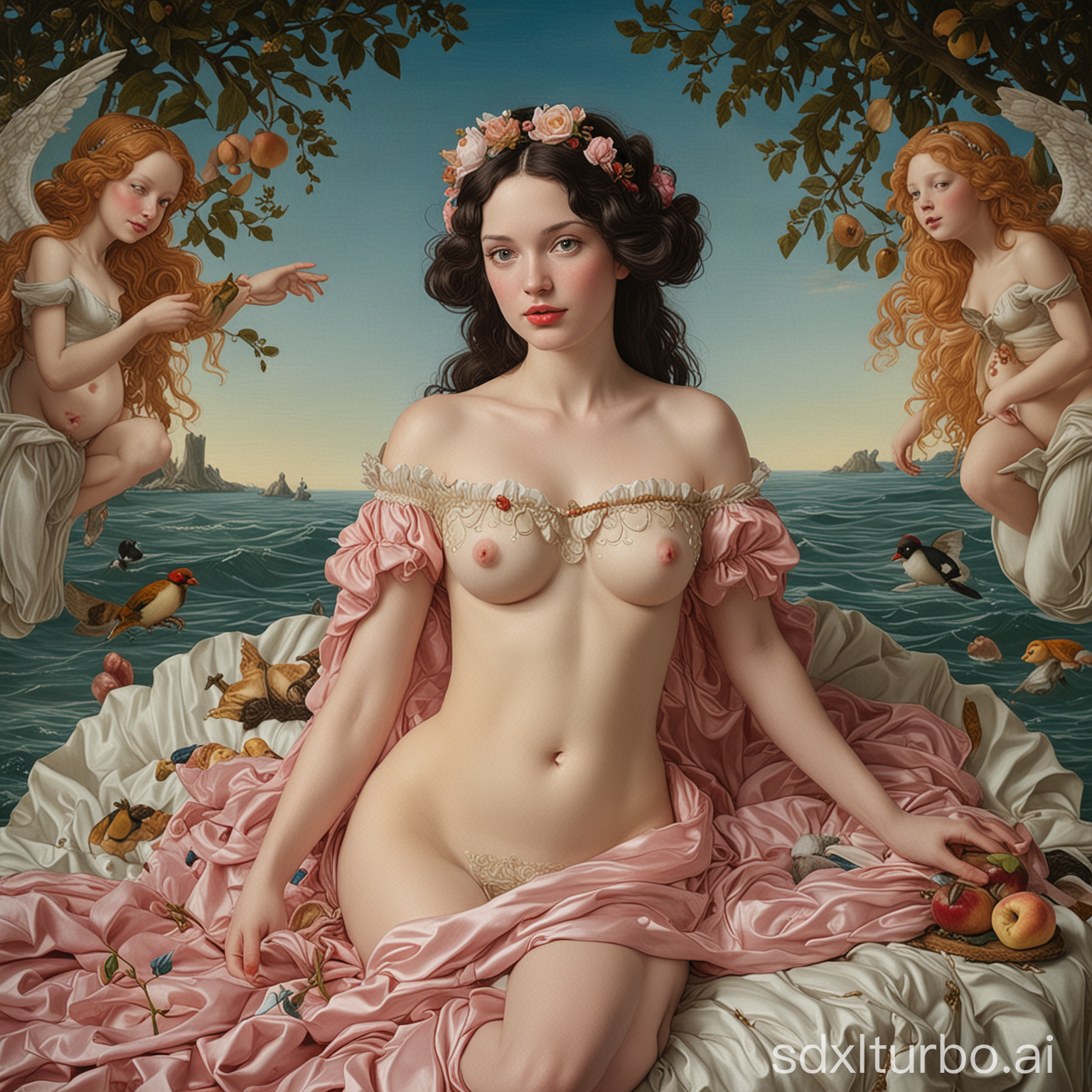} & \includegraphics[width=0.27\linewidth]{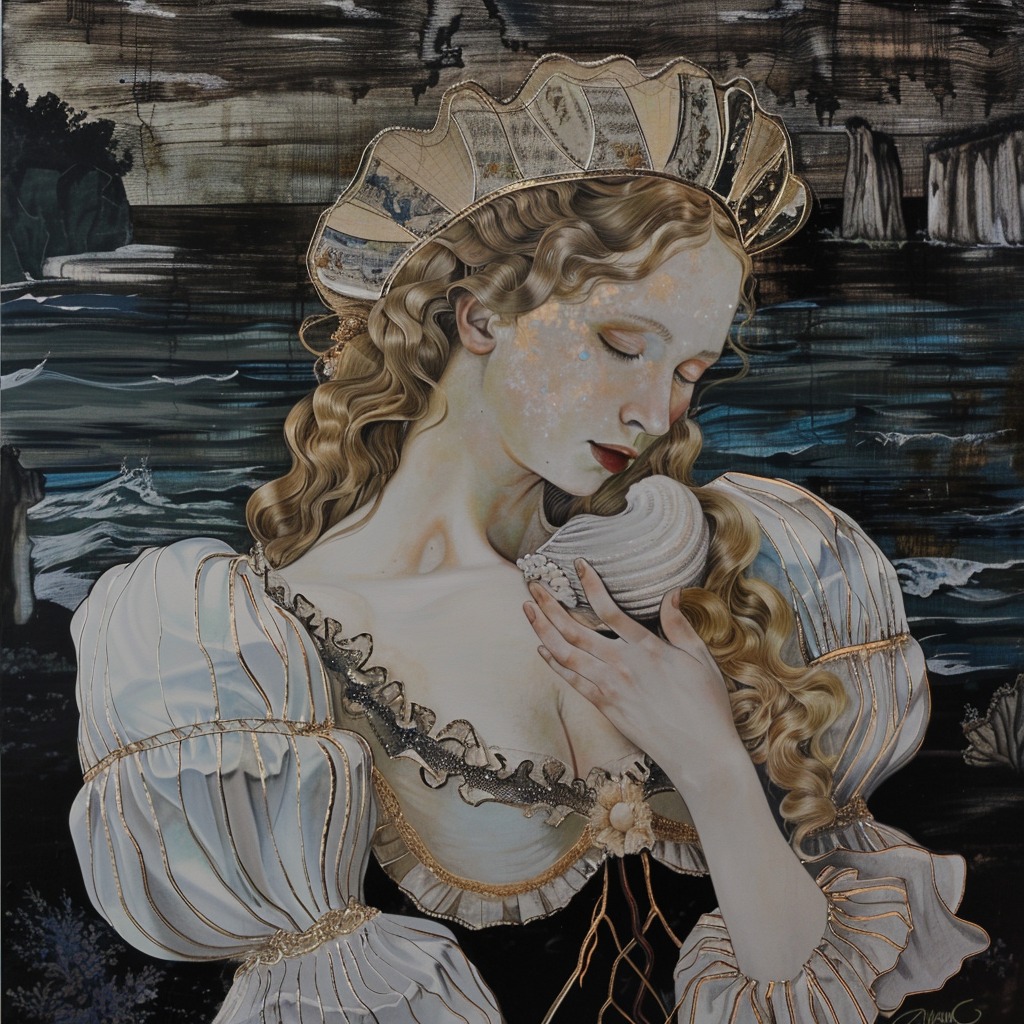} \\
    \end{tabular}
    \parbox{\textwidth}{\centering \emph{``A painting of snow white in the style of The Birth of Venus \\by Sandro Botticelli.''}}
    \vspace{-0.1cm}
    
    \begin{tabular}{cc}
        \includegraphics[width=0.27\linewidth]{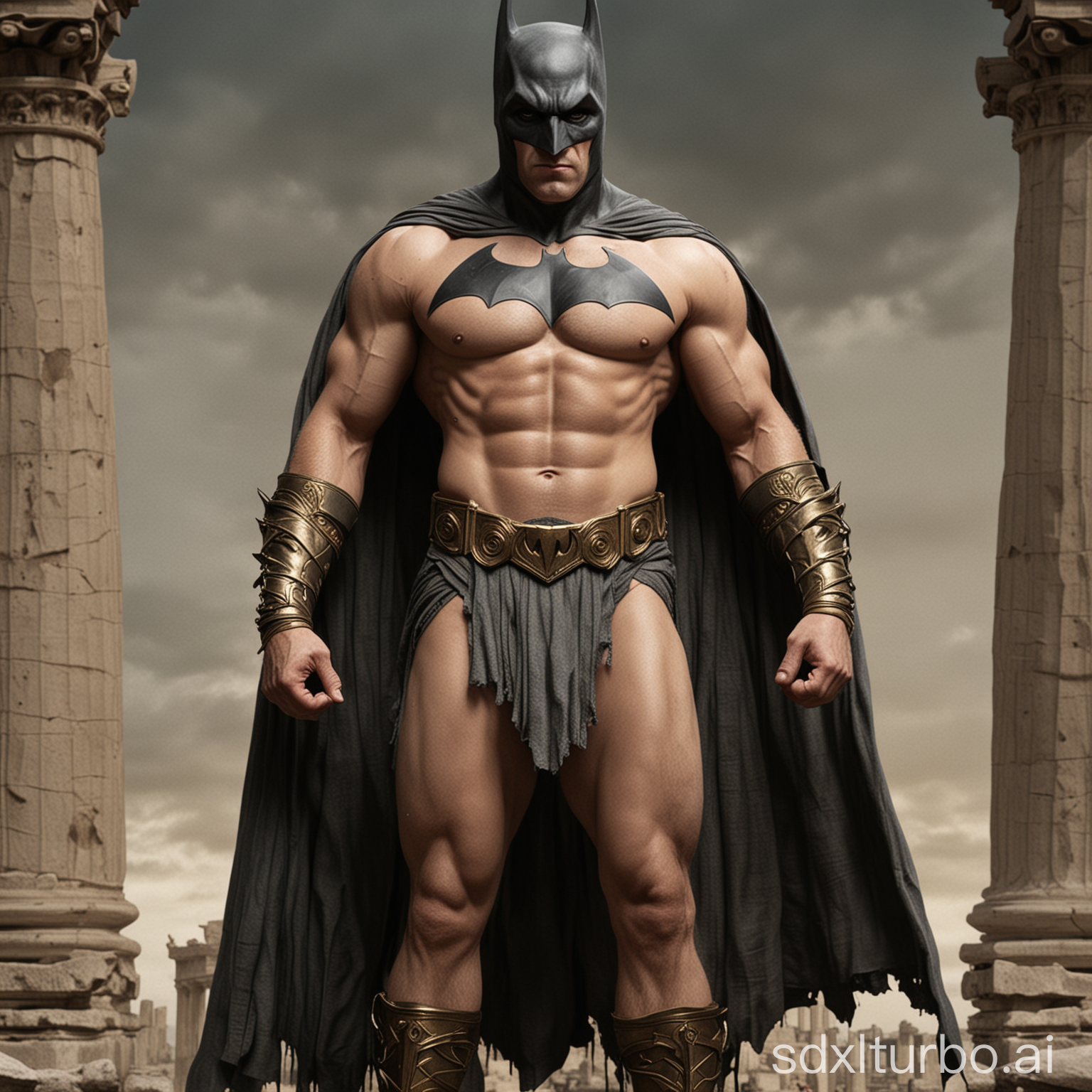} & \includegraphics[width=0.27\linewidth]{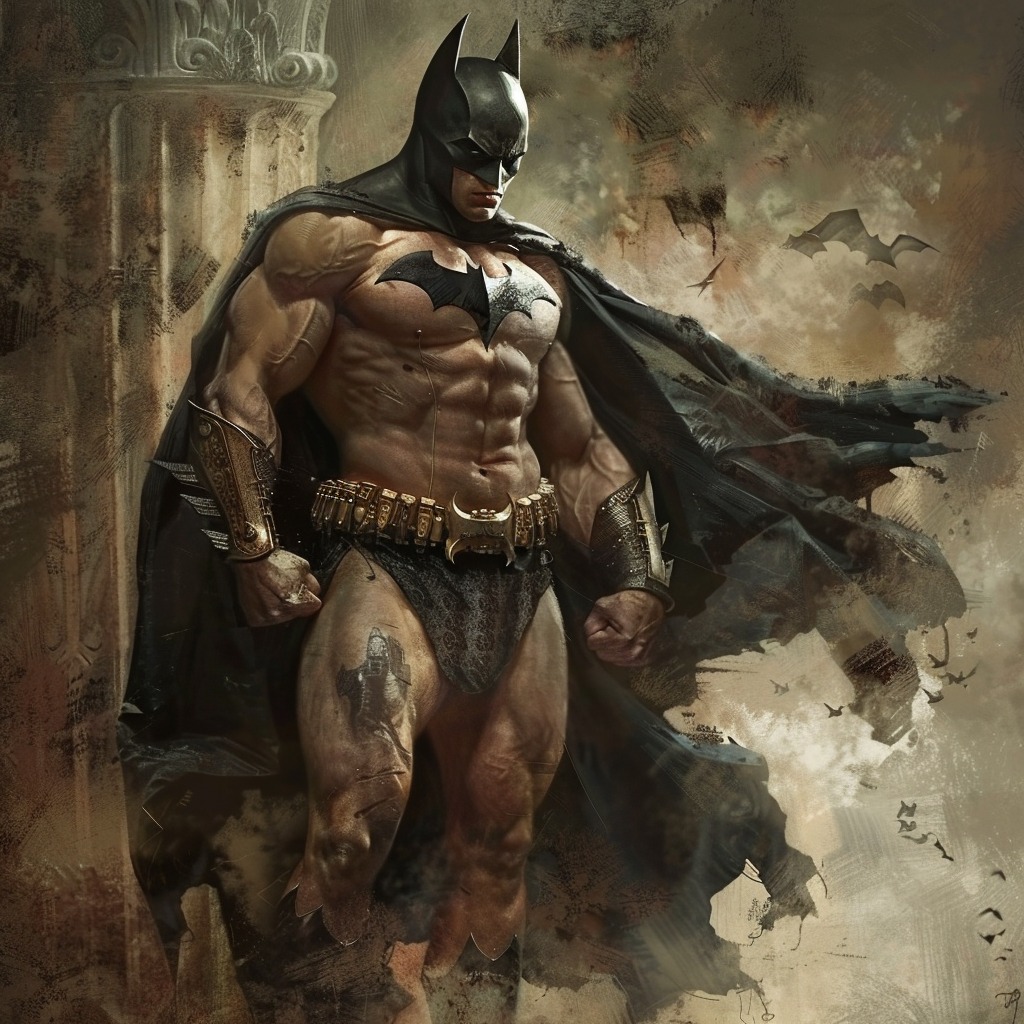} \\
        \includegraphics[width=0.27\linewidth]{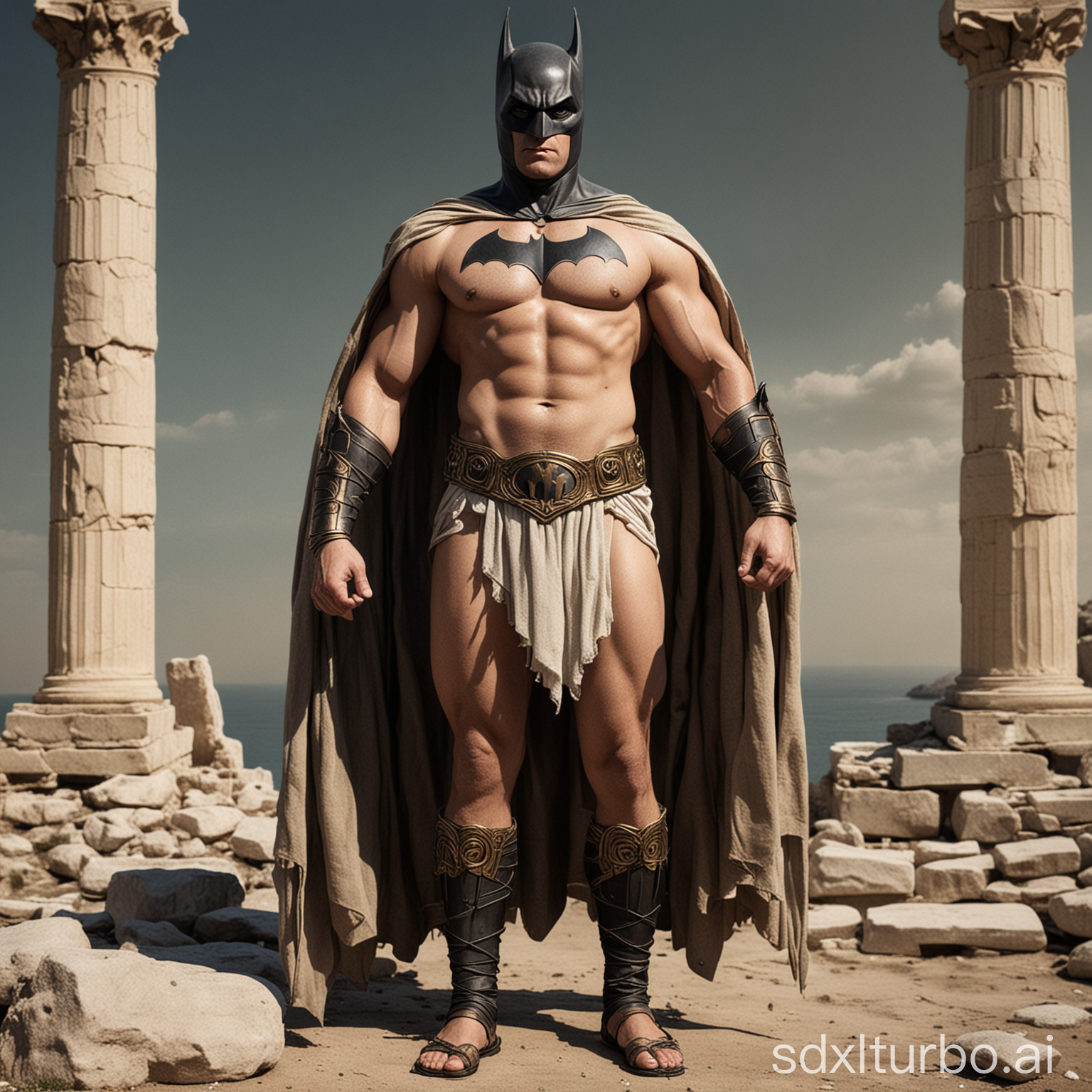} & \includegraphics[width=0.27\linewidth]{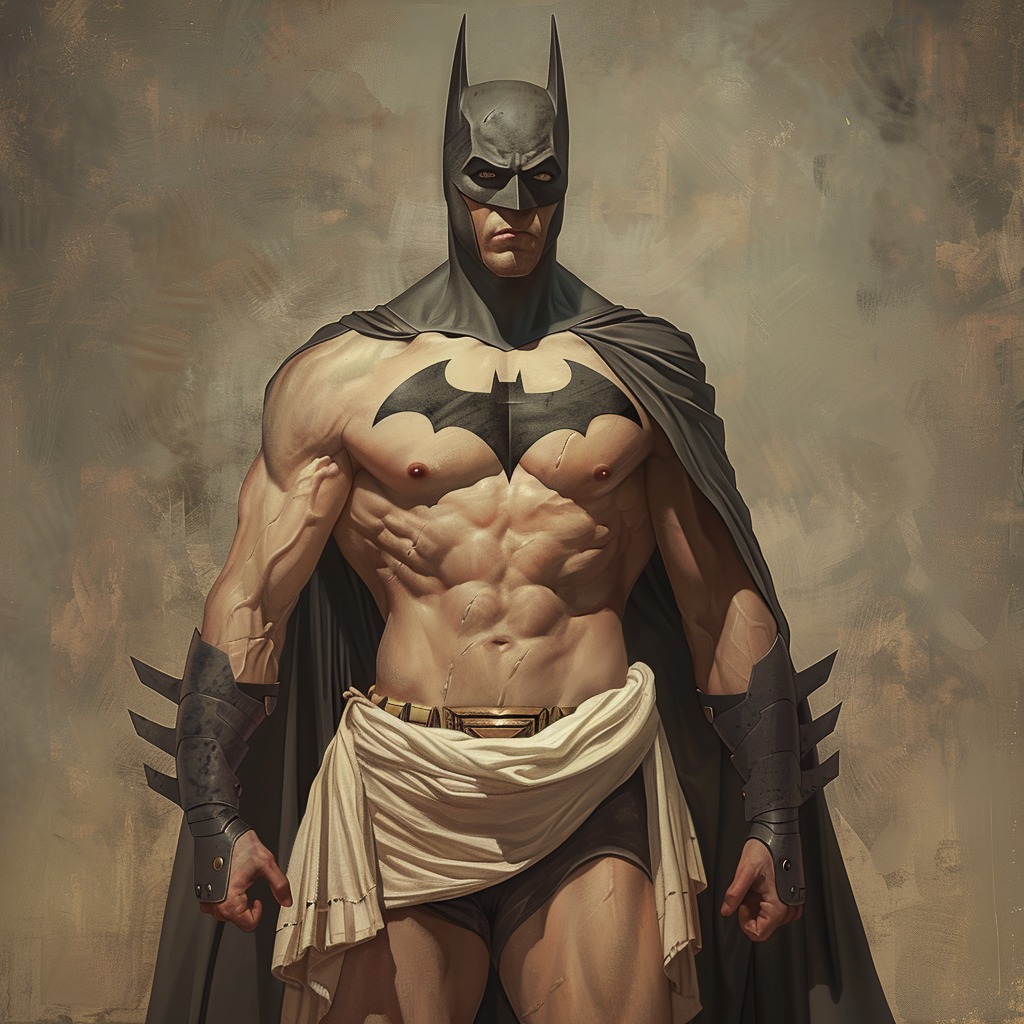} \\
    \end{tabular}
    \parbox{\textwidth}{\centering \emph{``Batman as a Greek god wearing a loincloth.''}}
    \vspace{-0.3cm}

    \caption{Example images generated by SDXL and Midjourney, showcasing the capability to create (partially or fully) nude images. \emph{(Left)} Images generated by SDXL. \emph{(Right)} Images generated by Midjourney. The images in the first and second rows were generated with the prompt \emph{“A painting of Snow White in the style of The Birth of Venus by Sandro Botticelli.”} The images in the third and fourth rows were generated with the prompt \emph{“Batman as a Greek God wearing a loincloth.”}}
    \label{ethics}
\end{figure*}

\section{Ethical concerns}\label{ethical_concerns}
The ability of generative models to create images that are indistinguishable from real human-made images comes with several ethical concerns. These so-called ‘Deepfakes’ have caused harm in many ways, including political misinformation, fraud through impersonation, and a broader loss of trust in digital media   \cite{mirsky2020, hancock2021, diakopoulos2021}. As a result, OpenAI and Google did not release their powerful text-to-image models fully open source. Users can only use filtered versions of their models through a web interface. These filtered versions have been trained on datasets that exclude potentially harmful images depicting violence, nudity or recognizable people (e.g., celebrities, political figures).

Stability AI opted for the transparent approach, releasing their models fully open source. This allows anyone to use and adjust their model for different purposes. Their models have been trained on less filtered datasets,  which include recognizable figures and artistic nudity such as nude paintings or sculptures. Fig.~\ref{ethics} illustrates how Stable Diffusion allows the generation of (semi-)nude images of recognizable figures. For illustration purposes, we opted for fictional characters and partially nude images. To make a comparison similar to the previous section, we also present images generated by Midjourney. Although we intended to include images generated by DALL-E 3, it did not permit us to do so due to content policy restrictions. However, it is possible to create harmful deepfakes using Stable diffusion and Midjourney. Fig.~\ref{ethics} shows that the Batman images are comparable in terms of nudity between SDXL and Midjourney. However, when generating Snow White in the style of \emph{The Birth of Venus}, there are noticeable differences in both styles and degrees of nudity. Specifically, in the case of Snow White, SDXL adheres more closely to the original composition, whereas Midjourney's generations align more with the original style. Additionally, SDXL produces significantly more explicit nudity compared to Midjourney in this example. That being said, Stability AI has instilled multiple safety barriers to avoid misuse by stating ethical and legal rules in Stable Diffusion’s permission license (e.g., stating you cannot spread nude or violent deepfakes) and including an AI-based safety classifier that filters the generated images. 

In the context of art, these Deepfakes raise numerous other concerns \cite{cejudo2023}. These models are often praised for their ability to replicate specific painters' style or art movements. The flipside of the coin is that this blurs the borders of copyright and raises questions regarding authorship of AI mediated art. These models allow anyone to generate images in the style of an artist, without proper copyright protection. \cite{shan2023} attempted to protect artists and their work by creating GLAZE, a model that allows artists to add an invisible watermark to their work that would distort a text-to-image model in learning their artistic style. As such, artists can protect their future work from copyright strikes. It is important to note that DALL-E 3 addresses concerns in the creative community by refusing requests to generate images in the style of living artists. It also maintains an updated blocklist of these artists' names. For a more thorough review on copyright infringement in relation to AI-generated art, we refer the reader to \cite{jiang2023}.

\section{Conclusions}\label{sec6}
Deep learning and its image processing applications are now at a totally different stage than they were a few short years ago. In the beginning of the last decade, it was groundbreaking that deep neural networks could classify natural images. Today, these models are capable of generating highly realistic and complex images based on simple text prompts. This allows individuals without programming knowledge to employ these powerful models. It is important to remember that the use of these models should be guided by ethical and responsible considerations. AI-generated art also invites reflection on broader issues, such as potential applications in art therapy and how creativity and authorship are understood. These tools are increasingly being explored by artists and they may influence artistic practice and discussions about the future of art. It is undeniable that AI is affecting authorship and copyright rules regarding art, which comes with many ethical considerations. As ChatGPT stated, \emph{“Some people believe that AI has the potential to revolutionize the way we create and think about art, while others are more skeptical and believe that true creativity and artistic expression can only come from human beings. Ultimately, the role of AI in the arts will depend on how it is used and the goals and values of the people who are using it.”}

\section*{Acknowledgements}

A-S. Maerten and D. Soydaner would like to thank ChatGPT for answering their questions patiently.

\noindent A-S. Maerten acknowledges support from the Research Foundation Flanders (FWO) through Grant No. 11C9524N.

\noindent D. Soydaner acknowledges support from the European Union through Grant No. 101053925 (ERC AdG, GRAPPA) awarded to Johan Wagemans.

\section*{Author contributions statement}

A-S. Maerten and D.Soydaner contributed equally to this work. All authors reviewed the manuscript.

\backmatter

\bibliography{sn-article}

\end{document}